\documentclass{article}

\makeatletter
\@ifundefined{ifanon}{\newif\ifanon}{}
\makeatother

\ifanon
  \usepackage[nonatbib]{neurips_2026}
\else
  \usepackage[preprint,nonatbib]{neurips_2026}
\fi

\usepackage[utf8]{inputenc}
\usepackage[T1]{fontenc}
\usepackage{hyperref}
\usepackage{url}
\usepackage{xcolor}
\usepackage{amsmath}
\usepackage{amssymb}
\usepackage{booktabs}
\usepackage{longtable}
\usepackage{array}
\usepackage{calc}
\providecommand{\real}[1]{#1}
\usepackage{caption}
\usepackage{enumitem}
\usepackage{verbatim}
\usepackage{graphicx}
\makeatletter
\def\maxwidth{\ifdim\Gin@nat@width>\linewidth\linewidth\else\Gin@nat@width\fi}
\def\maxheight{\ifdim\Gin@nat@height>\textheight\textheight\else\Gin@nat@height\fi}
\makeatother
\setkeys{Gin}{width=\maxwidth,height=\maxheight,keepaspectratio}

\usepackage{newunicodechar}
\newunicodechar{≈}{\ensuremath{\approx}}
\newunicodechar{≠}{\ensuremath{\neq}}
\newunicodechar{×}{\ensuremath{\times}}
\newunicodechar{−}{\ensuremath{-}}
\newunicodechar{∈}{\ensuremath{\in}}
\newunicodechar{·}{\ensuremath{\cdot}}
\newunicodechar{ρ}{\ensuremath{\rho}}
\newunicodechar{λ}{\ensuremath{\lambda}}
\newunicodechar{→}{\ensuremath{\rightarrow}}
\newunicodechar{←}{\ensuremath{\leftarrow}}
\newunicodechar{…}{\ensuremath{\ldots}}
\newunicodechar{‑}{-}
\newunicodechar{≥}{\ensuremath{\geq}}
\newunicodechar{≤}{\ensuremath{\leq}}
\newunicodechar{≪}{\ensuremath{\ll}}
\newunicodechar{¼}{\ensuremath{\sfrac{1}{4}}}
\newunicodechar{⅒}{\ensuremath{\sfrac{1}{10}}}
\newunicodechar{⁰}{\textsuperscript{0}}
\newunicodechar{¹}{\textsuperscript{1}}
\newunicodechar{²}{\textsuperscript{2}}
\newunicodechar{³}{\textsuperscript{3}}
\newunicodechar{⁴}{\textsuperscript{4}}
\newunicodechar{⁵}{\textsuperscript{5}}
\newunicodechar{⁶}{\textsuperscript{6}}
\newunicodechar{⁷}{\textsuperscript{7}}
\newunicodechar{⁸}{\textsuperscript{8}}
\newunicodechar{⁹}{\textsuperscript{9}}
\newunicodechar{§}{\S}
\newunicodechar{α}{\ensuremath{\alpha}}
\newunicodechar{β}{\ensuremath{\beta}}
\newunicodechar{γ}{\ensuremath{\gamma}}
\newunicodechar{δ}{\ensuremath{\delta}}
\newunicodechar{ε}{\ensuremath{\epsilon}}
\newunicodechar{σ}{\ensuremath{\sigma}}
\newunicodechar{μ}{\ensuremath{\mu}}
\newunicodechar{τ}{\ensuremath{\tau}}
\newunicodechar{φ}{\ensuremath{\phi}}
\newunicodechar{θ}{\ensuremath{\theta}}
\newunicodechar{ω}{\ensuremath{\omega}}
\newunicodechar{Σ}{\ensuremath{\Sigma}}
\newunicodechar{Δ}{\ensuremath{\Delta}}
\newunicodechar{Φ}{\ensuremath{\Phi}}
\newunicodechar{Ω}{\ensuremath{\Omega}}
\newunicodechar{Θ}{\ensuremath{\Theta}}
\newunicodechar{∇}{\ensuremath{\nabla}}
\newunicodechar{∂}{\ensuremath{\partial}}
\newunicodechar{∞}{\ensuremath{\infty}}
\newunicodechar{∘}{\ensuremath{\circ}}
\newunicodechar{⊆}{\ensuremath{\subseteq}}
\newunicodechar{⊂}{\ensuremath{\subset}}
\newunicodechar{±}{\ensuremath{\pm}}
\newunicodechar{‖}{\ensuremath{\Vert}}
\providecommand{\sfrac}[2]{#1/#2}

\makeatletter
\@ifundefined{c@none}{\newcounter{none}}{}
\makeatother

\providecommand{\tightlist}{\setlength{\itemsep}{0pt}\setlength{\parskip}{0pt}}
\makeatletter
\newsavebox\pandoc@box
\newcommand*\pandocbounded[1]{%
  \sbox\pandoc@box{#1}%
  \Gscale@div\@tempa\textheight{\dimexpr\ht\pandoc@box+\dp\pandoc@box\relax}%
  \Gscale@div\@tempb\linewidth{\wd\pandoc@box}%
  \ifdim\@tempb\p@<\@tempa\p@\let\@tempa\@tempb\fi%
  \ifdim\@tempa\p@<\p@\scalebox{\@tempa}{\usebox\pandoc@box}%
  \else\usebox\pandoc@box\fi%
}
\makeatother
\usepackage{framed}
\definecolor{shadecolor}{RGB}{248,248,248}

\hypersetup{
  colorlinks=true,
  linkcolor=blue,
  urlcolor=blue,
  citecolor=blue,
}

\title{The Reservoir Attention Network:\\
  Cross-Pass State in Pretrained Transformers\\
  via Content-Addressable Reservoir Injection}

\ifanon
  \author{Anonymous Authors}
\else
  \author{%
    Emma Leonhart\\
    \texttt{emma@topazcomputing.com}%
  }
\fi

\begin{document}
\maketitle

A feasibility and dynamics study of the Reservoir Attention Network (RAN), an architecture that
injects a fixed, randomly-initialized reservoir into the mid-layer attention of a pretrained
transformer to carry state across forward passes. Experiments span GPT-2 (124M, 355M) to
Qwen2.5 (0.5B, 1.5B) on a single consumer GPU. The tasks are minimal probes chosen to isolate
individual mechanisms; the broader always-alive agent vision is treated throughout as
compute-limited future work, not a claim of this paper. The reservoir is left \emph{untrained}
(fixed random) by design: this isolates whether untrained recurrent dynamics alone suffice to
carry usable cross-pass state, leaving trained recurrence as a complementary, more expensive
direction.

\hypertarget{abstract}{%
\section*{Abstract}\label{abstract}}
\addcontentsline{toc}{section}{Abstract}

Standard transformers are stateless across forward passes: no endogenous variable evolves
between calls, only position within the context window. We study whether a fixed,
randomly-initialized reservoir (in the echo-state-network sense), injected into the mid-layer
attention of a pretrained transformer and carried across passes, can endow the model with usable
state between calls without retraining the backbone, and identify the conditions under which the
injected state becomes usable signal. The study is conducted at GPT-2 to 1.5B scale on a single
consumer GPU.

We report four results. First, the \emph{injection mechanism} is decisive: writing the reservoir state
additively into the residual stream reproduces the known failure in which the model learns to
ignore the recurrent state (cross-context recall at chance, indistinguishable from a state-reset
baseline), whereas re-injecting the same state as content-addressable prefix pseudo-tokens that
upper layers attend to yields 100\% cross-context recall (1.00 versus a 0.17 reset-baseline,
reproducible). Second, the reservoir\textquotesingle s edge-of-chaos regime at spectral radius ≈ 1 persists under
real transformer activations, which over-drive a unit-scaled reservoir and require an input scaling
of approximately one-quarter to one-tenth. Third, cross-pass recall \emph{on the single
secret-word probe scales}: the apparent "GPT-2-small-only" boundary reported by prior single-machine
attempts was an undersized reservoir at an unmatched input scaling. Enlarging the reservoir to 2048
nodes and matching its input scaling to the model recovers probe recall across the Qwen family
(Qwen2.5-0.5B and 1.5B; stateful 0.83--1.00 versus a 0.17 control, reproduced across seeds), with
input scaling rather than parameter count as the decisive lever and a capacity ceiling of order tens
of items. This single-probe recall is a different measurement from the eight-task battery\textquotesingle s
\emph{symbolic content} recall (Result 4), which stays at the floor at 1.5B except under the retention
recipe; the two should not be conflated. The recovery is model-specific:
GPT-2-medium (355M) fails across a seven-point scaling sweep, so matched scaling is necessary but
not sufficient. Fourth, an eight-task stateful "battery" separates two cases: its temporal/agency
metrics are matched by a stateless ablation (not reservoir-driven), whereas its content recall \emph{can}
be retained under multi-task training once the stateless shortcut is denied its adapter capacity and
the learning rate is decayed: reservoir-driven recall then climbs monotonically to 1.00 against a
0.000 stateless control, with no collapse (a single run). We release weights and code. The
contribution is the injection-design finding, the dynamics characterization, and the scaling result,
with controlled negatives that bound them and a training recipe under which battery recall retains.

\hypertarget{contributions}{%
\section{Contributions}\label{contributions}}

\begin{itemize}
\tightlist
\item
  \textbf{Injection design decides whether carried state is usable.} Additive residual-stream injection
  reproduces the "learns to ignore the recurrent state" failure (recall at chance); content-addressable
  KV-prefix injection yields 100\% cross-context recall on GPT-2-small (1.00 vs a 0.17 wiped-reservoir
  control, reproducible).
\item
  \textbf{Reservoir dynamics characterized on real transformer activations.} The edge-of-chaos regime at
  spectral radius ≈ 1 persists, and real activations require an input scaling of approximately one-quarter to one-tenth to avoid
  saturation.
\item
  \textbf{Cross-pass recall (the secret-word probe) scales to a modern 1.5B model.} Sizing the reservoir up
  and matching its input scaling recovers single-probe recall across the Qwen family (0.83--1.00 vs 0.17
  control, reproduced); input scaling, not parameter count, is the decisive lever, and the prior
  "GPT-2-small-only" wall was an undersized reservoir. This is the probe, not the eight-task battery\textquotesingle s
  \emph{symbolic content} recall, which is a separate, harder setting that stays at the floor at 1.5B (see the
  controlled negatives below).
\item
  \textbf{Controlled negatives that bound the contribution.} A model-specific recovery boundary
  (GPT-2-medium fails across a seven-point scaling sweep, and 4-bit Hermes-3B does not converge), a
  capacity ceiling of order tens of items, and a stateless ablation showing the agentic battery\textquotesingle s
  temporal metrics are not reservoir-driven.
\end{itemize}

\hypertarget{research-question}{%
\section{Research Question}\label{research-question}}

Can a fixed, randomly-initialized reservoir injected into a pretrained transformer\textquotesingle s
mid-layer attention give the model explicit state between forward passes (a
time axis) without degrading its base capabilities, and what reservoir-dynamics
regime (spectral radius, reservoir size, injection depth) makes that injected state
usable signal rather than noise?

The Reservoir Attention Network (RAN) architecture introduces a fixed-random
recurrent substrate into the transformer\textquotesingle s attention mechanism. We refer to a
specific instantiation of this architecture as a Reservoir Agent.

We scope the question as a feasibility and dynamics study at small scale
(GPT-2-scale base, single machine). The full vision (forking an agent harness into an
always-alive runtime and N-seed LoRA selection at agent scale) is the long-horizon
target, outside the scope of this study.

\hypertarget{scope-and-claims}{%
\section{Scope and Claims}\label{scope-and-claims}}

To be explicit about the boundary of the claims:

\begin{itemize}
\tightlist
\item
  \textbf{The tasks are minimal mechanism-isolating probes, not agentic demonstrations.}
  Secret-word recall and the trigger-based silence policy are intentionally the
  \emph{simplest} tasks that a stateless model structurally cannot do; their job is to
  isolate one variable (does carried state become usable signal, and under which
  injection design), not to exhibit complex agentic reasoning. We make no claim of
  complex agentic behaviour at this scale; that is named as future work, not shown here.
\item
  \textbf{The complexity-theory argument is motivation, not a result.} The TC⁰ / FO(M)
  framing explains \emph{why} cross-pass state is the interesting lever; we state plainly that
  there is no proof a finite-precision reservoir lifts the per-pass bound, and we
  treat it as the central open theoretical question, not an established finding.
\item
  \textbf{The GPT-2-medium / 4-bit-3B negatives and the KV-append integration constraint are
  limitations, stated as such.} The cross-pass recall result holds at GPT-2-small and across the
  Qwen family (0.5B, 1.5B) with model-matched input scaling, but not at GPT-2-medium (chance
  across a seven-point scaling sweep) or 4-bit Hermes-3B (injection verified as correctly wired but
  non-converging; 4-bit is a confound and a clean bf16 3B test does not fit this GPU). The injection
  variant that runs (KV-prefix) prepends the reservoir state as attendable pseudo-tokens; the richer
  strict mid-layer KV-append (appending reservoir rows directly into a layer\textquotesingle s K/V) is not wired into
  the live model because HuggingFace\textquotesingle s \texttt{generate} exposes no hook for appending external KV entries, so
  our results use a bespoke forward loop with the KV-prefix path; this is an integration constraint, not
  a difference in the conclusion, and the implementation is open. Neither limitation is hidden; both
  bound the contribution.
\item
  \textbf{The contribution is the injection-design finding.} What this study \emph{does}
  establish, decisively and reproducibly on GPT-2, is that how the reservoir is
  injected is the deciding factor: additive injection is ignored (chance recall), while
  content-addressable KV-prefix injection gives 100\% cross-context recall. That negative-
  then-positive result is the central contribution.
\end{itemize}

\hypertarget{architecture}{%
\section{Architecture}\label{architecture}}

Every forward pass is one reservoir tick. At a mid-depth injection layer Lk, attention
runs jointly over the token hidden states and a set of reservoir nodes (extra
keys/values). The reservoir reads the layer\textquotesingle s attention output through a fixed random
projection W\_in and writes its state back through a learned readout W\_out (both at the
same layer, every pass), so the reservoir state accumulates a history of the model\textquotesingle s
own attention dynamics across passes. The reservoir update is

r(t) = (1 − a)·r(t−1) + a·tanh( W\_r · r(t−1) + W\_in · x(t) )

a leaky-integrator update with leak rate a ∈ (0, 1{]} (the plain form r(t) = tanh(·) is
the special case a = 1; this leaky form is the one used in the dynamics sweeps), with
W\_r a fixed random sparse matrix scaled to a target spectral radius, W\_in fixed random,
and W\_out (plus light upper-layer LoRA) the only trained parameters. The leak rate and
reservoir size are the two levers for how much cross-pass state is carried. The lower
layers are frozen. Because the reservoir state is decoupled from the context window, it
persists across genuinely independent forward passes, including unprompted ticks.

The contrast with a standard transformer is the essential contrast. In a standard transformer
(below), every token attends to every other token within a single forward pass, and
nothing endogenous survives once that pass ends: the only memory is position inside the
context window, which the architecture is free to wipe.

\begin{figure}
\centering
\includegraphics{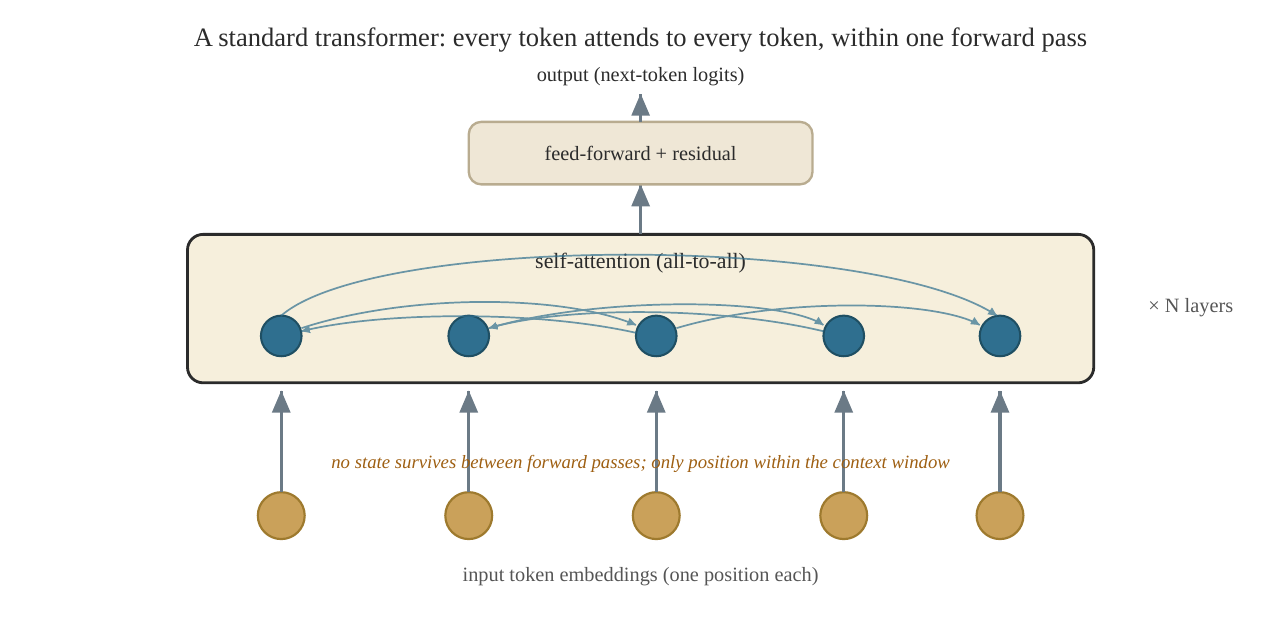}
\caption{A standard transformer: within one forward pass every token attends to every token, then the pass ends. No endogenous state survives between passes: the only memory is position inside the context window.}
\end{figure}

The RAN keeps that backbone intact and grafts a single memory channel onto one mid-depth
attention layer. The reservoir sits beside the model as a fixed recurrent pool; it reads
the layer\textquotesingle s attention activations through the fixed projection W\_in, updates its own state
r(t), and writes that state back as attendable key/value prefix nodes through the trained
readout W\_out. The recurrent state r(t) → r(t+1) is the part that survives a context wipe:
the explicit time axis the standard diagram lacks.

\begin{figure}
\centering
\includegraphics{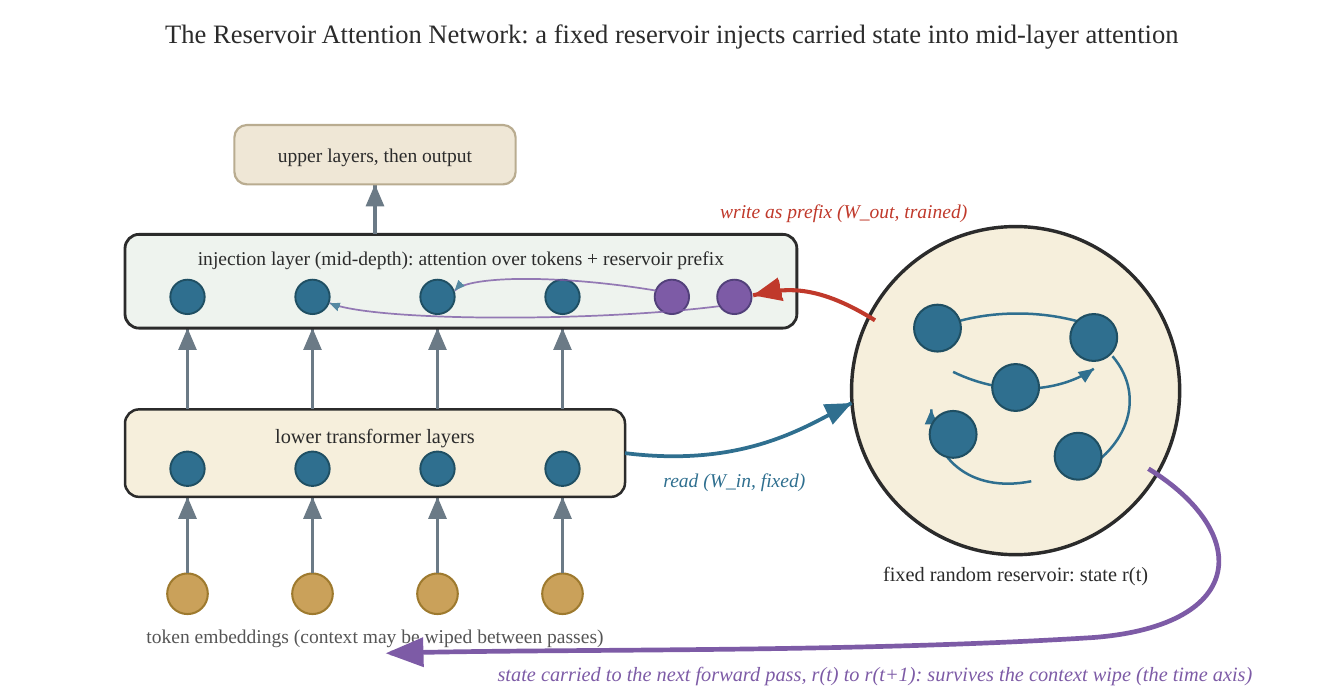}
\caption{The Reservoir Attention Network: at a mid-depth injection layer the attention runs jointly over the token nodes and reservoir prefix nodes. A fixed random reservoir reads the layer activations through W\_in, evolves its state, and writes back as a learned prefix (W\_out); its state r(t) is carried to the next forward pass, surviving the context wipe.}
\end{figure}

\begin{figure}
\centering
\includegraphics{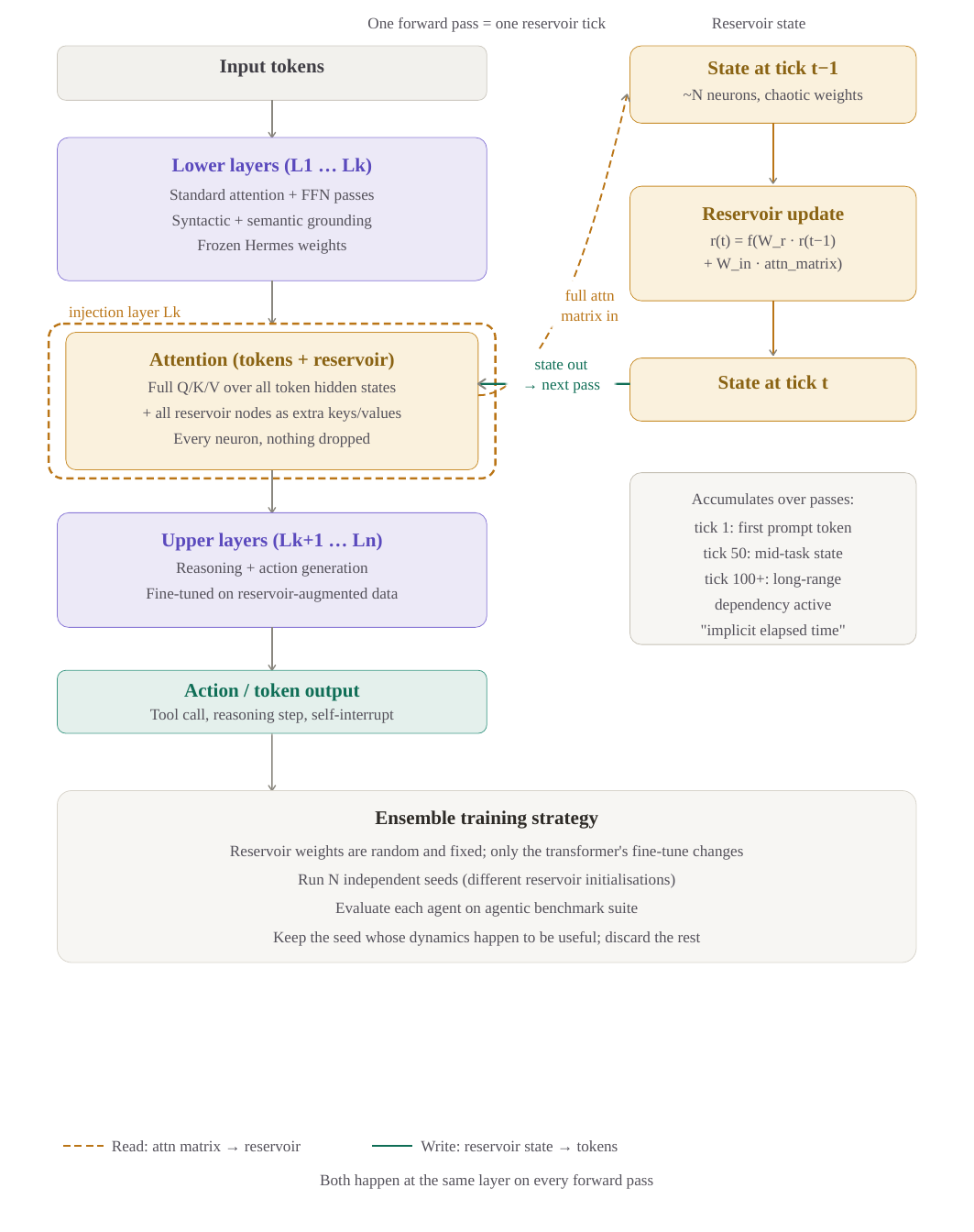}
\caption{Architecture: a fixed random reservoir reads the attention output at a mid-depth layer through W\_in, updates its state r(t), and writes back through the learned readout W\_out, so the reservoir carries a history of the model\textquotesingle s own attention dynamics across forward passes.}
\end{figure}

\begin{figure}
\centering
\includegraphics{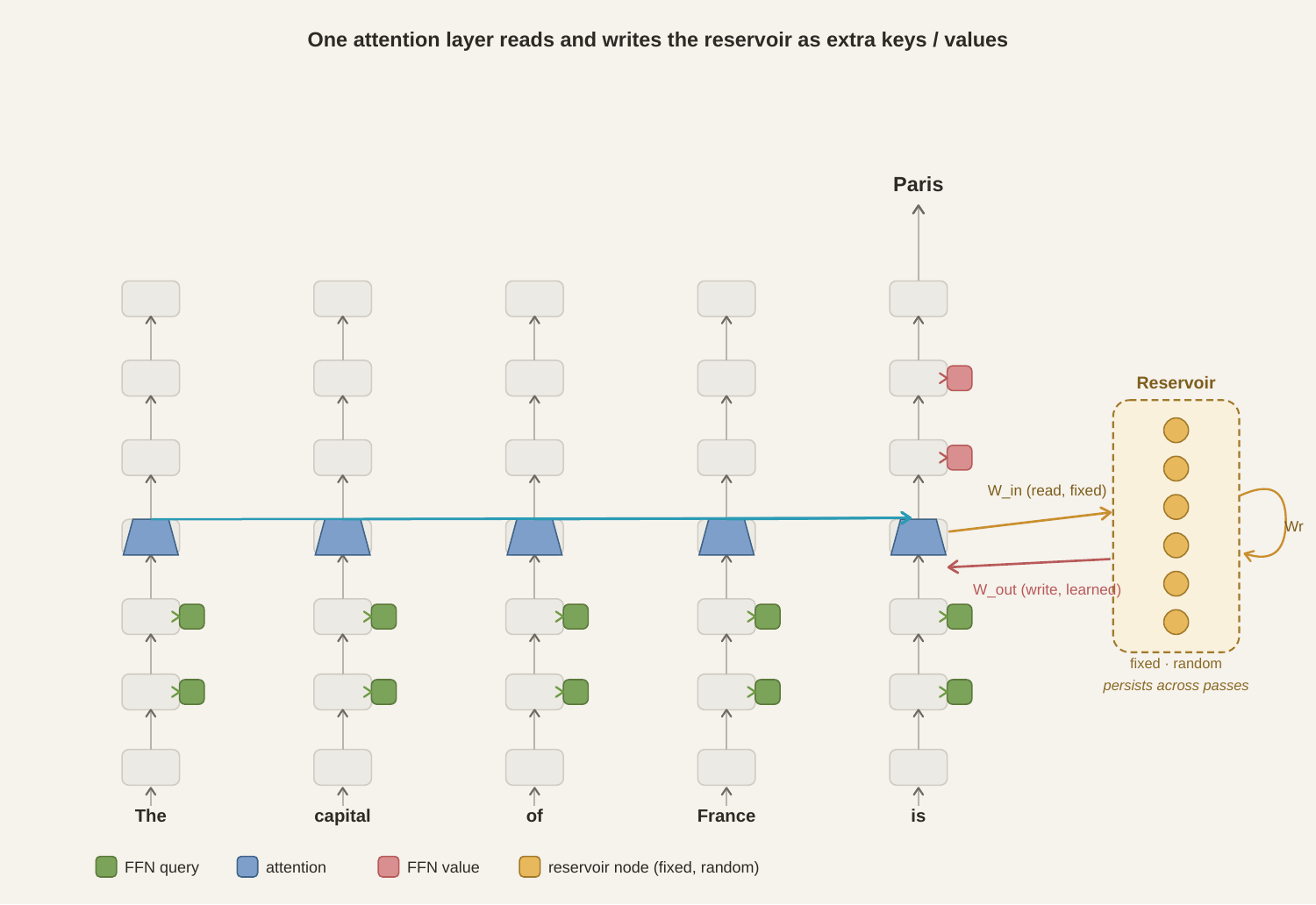}
\caption{The two injection variants. Additive: the readout is added as a single bias into the residual stream (the model learns to ignore it). Content-addressable (KV-prefix): the reservoir state becomes attendable key/value pseudo-tokens the upper layers query, the variant that yields cross-pass recall.}
\end{figure}

Design rationale: the reservoir adds memory to a proven system, rather than being the system.
Classical reservoir computing has a well-known reliability problem: the quality of a fixed random
reservoir varies, so getting a usable one typically means generating and selecting over many
candidates, and the readout carries all task performance. This project deliberately sidesteps that:
the reservoir is injected into a \emph{pretrained, well-proven} transformer and the model is \emph{fine-tuned}
(readout + light LoRA) to read from it, so the reservoir\textquotesingle s only job is to add a memory channel
to a system that already works, not to be the computational substrate. This is consistent with our
N-seed selection experiment, which finds that \emph{which} fixed reservoir one draws is not a significant
predictor of performance at our budget (selection is dominated by run-to-run training noise); so
relying on reservoir selection would be fragile, and folding the burden onto fine-tuning the readout
into a capable backbone is the more robust design.

\hypertarget{related-work}{%
\section{Related Work}\label{related-work}}

This section reviews the prior literature, grounding the work in three bodies of
prior art. Full citations are in the References.

Reservoir computing. The defining move, fixing the recurrent weights (W\_r, W\_in) at
random and training only a readout, is the echo-state-network (Jaeger, 2001) /
liquid-state-machine (Maass, 2002) paradigm, surveyed by Lukoševičius and Jaeger (2009).
Usable behaviour is governed by the \emph{echo state property} (the influence of past state
and input must fade); scaling the recurrent matrix to spectral radius ρ \textless{} 1 \emph{almost
always} secures it, but ρ \textless{} 1 is neither exactly necessary nor sufficient, and the
"operate at the edge of chaos" heuristic is disputed: short-term memory capacity can
peak away from the edge. Reservoir memory is fading transient memory, capacity-bounded
by reservoir size (linear memory capacity ≤ N). So the classical recipes are a \emph{prior,
not an answer} for our regime, which is exactly what our dynamics sweep measures, and
reservoir size and leak rate are the controls for how much cross-pass state is carried.

\begin{figure}
\centering
\includegraphics{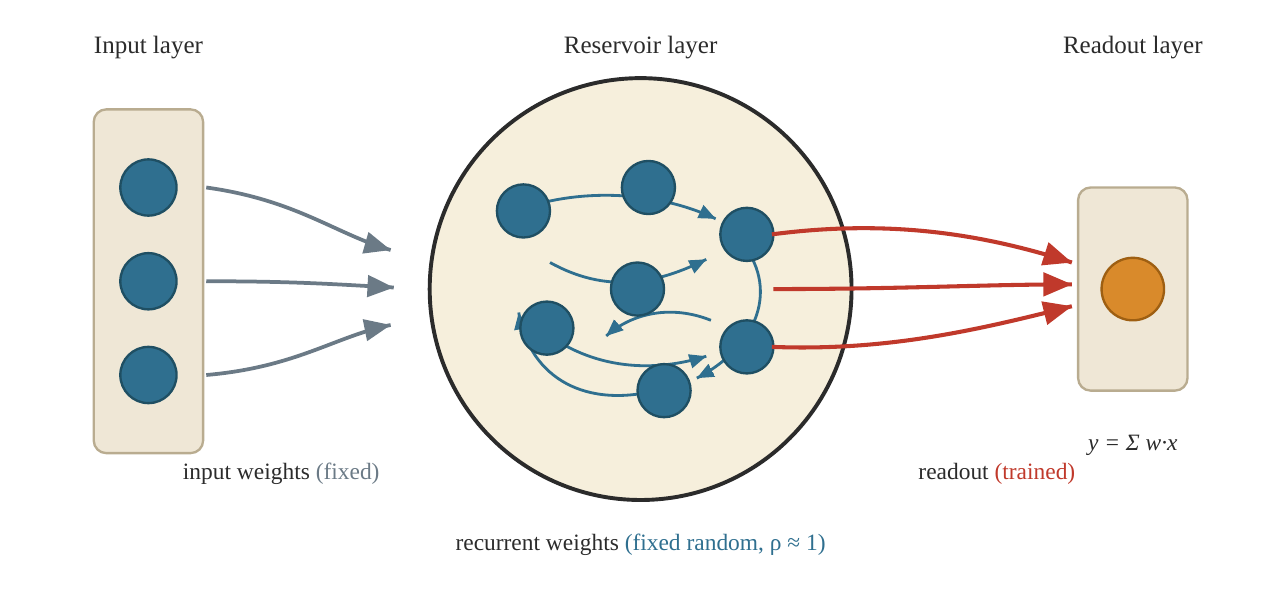}
\caption{Reservoir computing in the classical form: an input is projected by a fixed random W\_in into a fixed random recurrent pool (the reservoir, spectral radius ρ ≈ 1), and only a linear readout W\_out is trained. The RAN relocates this paradigm: the reservoir reads and writes a pretrained transformer\textquotesingle s mid-layer attention instead of a scalar input, and attention is the readout surface.}
\end{figure}

\begin{figure}
\centering
\includegraphics{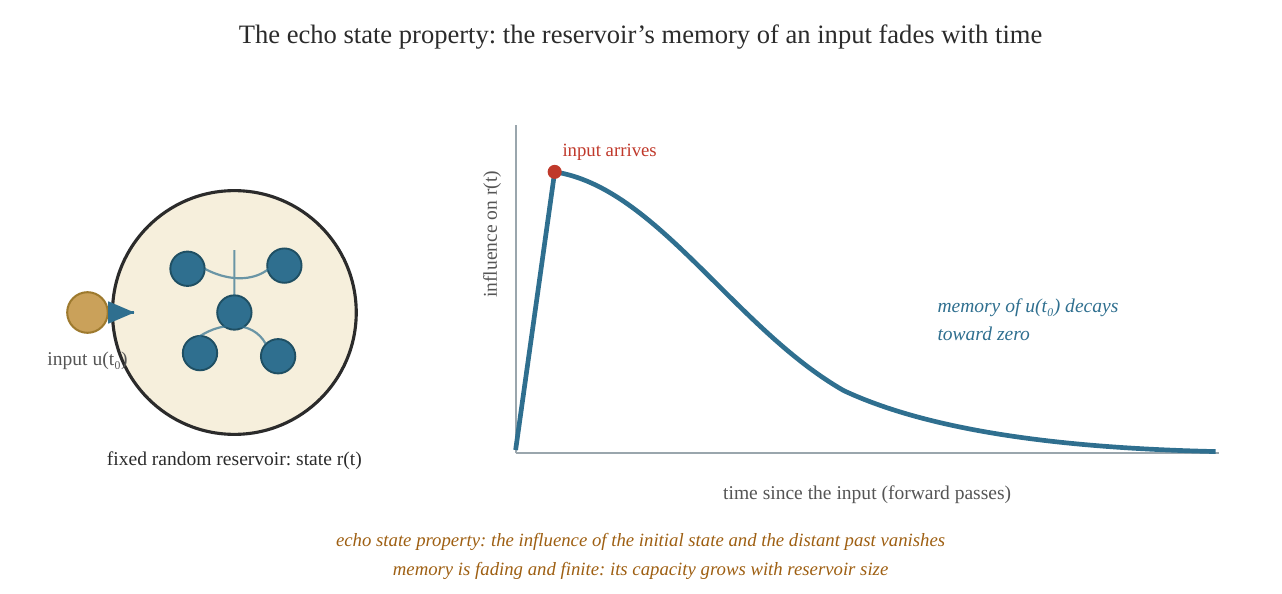}
\caption{The echo state property, the condition that makes a reservoir usable: after an input arrives, its influence on the reservoir state decays toward zero over subsequent passes, and the influence of the initial state and the distant past vanishes. Reservoir memory is therefore fading and finite (its linear memory capacity grows with reservoir size, ≤ N), which is why reservoir size and leak rate are the controls for how much cross-pass state is carried.}
\end{figure}

The stateless-transformer ceiling. A fixed-depth, finite-precision transformer is,
per forward pass, confined to a low complexity class: saturated/float transformers are in
TC⁰ (Merrill, Sabharwal \& Smith, 2022), and log-precision transformers are captured by
first-order logic with majority quantifiers, FO(M) (Merrill \& Sabharwal, 2023); fixed-size
attention cannot model unbounded hierarchical structure without growing depth (Hahn,
2020). Cross-pass state is the documented lever past this ceiling (the upper-bound proof
explicitly breaks under input feedback), but the known escapes (Pérez et al., 2019;
Siegelmann \& Sontag, 1995) require \emph{arbitrary precision}. We therefore pose whether a
finite-precision reservoir state lifts the bound as an open question, not a result.

Recurrence-augmented transformers: the closest prior art, and the gap. A decade of
work adds recurrence, state, or memory to transformers. Classified on the two axes that
matter here (is the recurrence \emph{trained} or \emph{fixed-random}, and does state persist
\emph{within a sequence} or \emph{across independent passes}):

\begin{longtable}[]{@{}
  >{\raggedright\arraybackslash}p{(\columnwidth - 4\tabcolsep) * \real{0.3333}}
  >{\raggedright\arraybackslash}p{(\columnwidth - 4\tabcolsep) * \real{0.3333}}
  >{\raggedright\arraybackslash}p{(\columnwidth - 4\tabcolsep) * \real{0.3333}}@{}}
\toprule\noalign{}
\begin{minipage}[b]{\linewidth}\raggedright
System
\end{minipage} & \begin{minipage}[b]{\linewidth}\raggedright
Recurrence
\end{minipage} & \begin{minipage}[b]{\linewidth}\raggedright
State persistence
\end{minipage} \\
\midrule\noalign{}
\endhead
\bottomrule\noalign{}
\endlastfoot
Transformer-XL, Compressive Transformer & trained & within sequence (cached segment) \\
Universal Transformer & trained & intra-pass depth (not temporal) \\
Block-Recurrent Transformer & trained (gates) & within sequence \\
Memorizing Transformers & trained retrieval & within document (stored kNN) \\
Recurrent Memory Transformer & trained (memory tokens) & across segments of one sequence \\
RWKV, RetNet, S4 / Mamba & trained & within sequence (RNN/SSM form) \\
Titans & trained (test-time updates) & within a stream \\
RAN (this work) & fixed-random & across independent forward passes \\
\end{longtable}

Every prior system uses \emph{trained} recurrence carrying state \emph{within} a sequence or
segment chain. The RAN occupies the empty cell: a \emph{fixed-random} reservoir whose state
persists across genuinely independent passes (including unprompted, no-input ticks),
injected into a \emph{pretrained, frozen} backbone and trained only through a readout plus
light LoRA. Block-Recurrent Transformer independently documents the failure mode we
observe directly (§ "the model learns to ignore the recurrent state"): tasks must
structurally require the carried state or it is ignored. The nearest recent items,
Reservoir Transformers (Shen et al., 2021, arXiv:2012.15045), Echo State Transformer
(Bendi-Ouis \& Hinaut, 2025, arXiv:2507.02917), Echo Flow Networks (Liu \& Xu, 2025,
arXiv:2509.24122), and Frozen-in-Time (Singh et al., 2025, arXiv:2508.18130), each differ on at
least one load-bearing axis (none
injects into a \emph{pretrained} LLM\textquotesingle s attention with a \emph{fixed-random} reservoir carrying
state across \emph{independent} passes); they are trained-from-scratch and within-sequence.
A concurrent line compares reservoir computing \emph{directly against} transformers as language models
(Köster \& Uchida, 2025, arXiv:2507.15779), finding reservoirs more compute-efficient but lower in
prediction quality; this is orthogonal to our aim, which is to \emph{inject} a reservoir into a pretrained
transformer to add cross-pass memory, not to replace the transformer with a reservoir.
(The 2025 items are recent preprints; arXiv identifiers are given so they can be verified.)
This places the combination as novel against the verified prior art, with the caveat that
the recent landscape moves quickly and a verified absence is not a proof of absence.

\hypertarget{motivation-complexity-theoretic-framing}{%
\section{Motivation: Complexity-Theoretic Framing}\label{motivation-complexity-theoretic-framing}}

Three framing points, stated at the level of \emph{kind} of capability, not level of capability:
motivation for the design, not results. Grounding and citations are given in the References.

\textbf{1 · An explicit time dimension.} A standard transformer represents time as token
\emph{position}: an index into a sequence, not a dimension the model evolves along. With
the reservoir, the state r(t) evolves continuously across forward passes:
r(t) = (1−a)·r(t−1) + a·tanh(W\_r·r(t−1) + W\_in·x(t)), so r at pass N is causally
downstream of every pass since t=0. This is not positional encoding and not context
length; both reset or slide with the input. The reservoir state is decoupled from the
context window (it survives context truncation), which is precisely what a "time axis"
means here: an endogenous variable the model accumulates along, independent of the
input sequence.

\textbf{2 · The expressivity motivation (one sentence; no result claimed).} A finite-precision
transformer is bounded per forward pass to a low complexity class (TC⁰/FO(M)) and cross-pass
state is the standard lever past it; this only motivates \emph{why} cross-pass state is interesting,
and we prove no separation and nothing below depends on it (full discussion and citations in the
References).

\textbf{3 · Endogenous state (a bounded structural analogy).} The reservoir introduces
endogenous state that evolves independently of external input. This is a structural
property, not a capability claim: a stateless transformer carries no variable that
persists across independent passes, whereas this architecture does. That persistent
variable is a \emph{precondition} for the simple stateful behaviours this paper probes
(noticing an unresolved thread, estimating elapsed time, self-initiating), which a
stateless model cannot represent regardless of its capability level. We make no claim
about general intelligence, and we draw no analogy beyond this structural one.

\hypertarget{method}{%
\section{Method}\label{method}}

\begin{enumerate}
\def\labelenumi{\arabic{enumi}.}
\tightlist
\item
  Reservoir core. A tested echo-state reservoir with spectral-radius control and
  dynamics observability (variance, saturation fraction, effective rank, trajectory
  distinguishability).
\item
  Dynamics characterization. Drive the reservoir across a grid of spectral radius
  and size; locate the regime where the state is non-saturating, non-exploding, and
  carries distinguishable trajectories across input histories (H2), and test whether
  the optimum sits at the classical edge-of-chaos prior (which the literature reports
  is disputed).
\item
  Model surgery (H1). Inject the reservoir into a mid layer of GPT-2-small and
  verify that, with the readout zeroed, the base model\textquotesingle s outputs are unchanged;
  i.e. the architecture degrades gracefully to vanilla behaviour.
\end{enumerate}

\hypertarget{results}{%
\section{Results}\label{results}}

Figures supporting the results are collected in Appendix: Figures.

\hypertarget{h1-the-reservoir-injects-without-breaking-the-base-model}{%
\subsection{H1: the reservoir injects without breaking the base model}\label{h1-the-reservoir-injects-without-breaking-the-base-model}}

Hooking a mid-depth block of pretrained GPT-2 so the block\textquotesingle s hidden states drive the
reservoir and its state is written back into the residual stream (\texttt{h\textquotesingle{}\ =\ h\ +\ W\_out·r(t)}):

\begin{itemize}
\tightlist
\item
  \textbf{Non-destruction (a wiring sanity check, not a finding).} With the readout \texttt{W\_out\ =\ 0}
  the injected model\textquotesingle s next-token logits are \emph{identical} to vanilla GPT-2 (\texttt{allclose},
  atol 1e-5). This is trivially true by construction; we report it only as a wiring check: with the readout zeroed,
  the injected path reduces to the identity, which confirms the hook is placed at the intended
  layer and the computation graph is intact.
\item
  \textbf{The injection is live.} A nonzero \texttt{W\_out} changes the logits, and the reservoir
  state after two forward passes differs from after one, a genuine cross-pass time
  axis.
\end{itemize}

\hypertarget{h3-a-trained-readout-extracts-history-a-stateless-model-cannot}{%
\subsection{H3: a trained readout extracts history a stateless model cannot}\label{h3-a-trained-readout-extracts-history-a-stateless-model-cannot}}

On the delay-memory task (drive the reservoir with i.i.d. input u(t); train a linear
ridge readout to reproduce u(t−τ)), the readout on the reservoir state recovers the
input from \textasciitilde18 steps back at R² \textgreater{} 0.5 and \textasciitilde12 steps back at R² ≈ 1, with a total
linear memory capacity of 17.4 (Σ R² over τ ≥ 1). The stateless baseline,
the same readout trained on the \emph{current} input u(t), scores exactly 0 at every
delay ≥ 1, because i.i.d. inputs carry no information about their own past. So the
information needed to answer is provably \emph{in the carried state, not the input}: a light
trained readout makes the reservoir\textquotesingle s history usable, and a stateless model structurally
cannot match it. This is the H3
mechanism on a clean synthetic task; doing it on a \emph{semantic} agent task (unresolved
thread, elapsed time) is future work that needs the readout trained through the LM.

\hypertarget{n-seed-selection-the-mechanism-works-the-low-cost-pre-selection-proxy-does-not}{%
\subsection{N-seed selection: the mechanism works; the low-cost pre-selection proxy does not}\label{n-seed-selection-the-mechanism-works-the-low-cost-pre-selection-proxy-does-not}}

Running the plan\textquotesingle s N-seed selection at small scale (train each of 12 fixed reservoir
seeds\textquotesingle{} readout on the delay-memory task, rank by memory capacity, keep the best): the
seeds genuinely differ (memory capacity ranges 17.4 to 20.7, a \textasciitilde19\% spread), so the
selection is worth doing. But the open "seed pre-selection proxy" question (can a low-cost
\emph{untrained} dynamics metric predict which seed trains best, to skip training?) gets a
clean negative answer for this proxy: the untrained participation ratio has no
rank correlation with trained memory capacity (Spearman ρ = 0.08, p = 0.80, n=12).
So seeds cannot be pre-filtered by participation ratio; the N-seed \emph{training} does real
work this dynamics proxy can\textquotesingle t shortcut. The cost implication,
stated plainly: because this proxy fails, selecting a good fixed reservoir
currently requires training each seed\textquotesingle s readout, i.e. genuine trial-and-error, not a
low-cost pre-filter. Finding an untrained proxy that \emph{does} correlate is open work; until
then the selection cost scales with the number of seeds tried.

Per-seed recall spreads widely, but at this budget it is dominated by training noise,
not cleanly by reservoir quality (a correction). Training a population of fixed reservoir
seeds end-to-end on the cross-pass task (GPT-2, 250 steps each) gives recall from 1.00 to
chance (0.17) across seeds (populations of 12 and 20 are published at
\texttt{EmmaLeonhart/reservoir-agent-gpt2-batch-n12} and \texttt{-n20}). It is tempting to read that
spread as reservoir \emph{quality}, but the two runs share seed indices, which gives a natural
replication, and it does not hold up: the same seed (identical fixed reservoir, same
setting) lands at very different recall across the two runs (e.g. seed 0 at 0.33 vs 1.00,
seed 1 at 1.00 vs 0.33), with mean \textbar Δrecall\textbar{} ≈ 0.47 over the 12 shared seeds, nearly as
large as the full spread. So at 250 steps the outcome is run-to-run noise-dominated
(CUDA non-determinism + an under-trained regime + the trainable readout/LoRA init not being
seeded by the reservoir seed), and a single run per seed cannot separate reservoir quality
from training noise. Consistently, no untrained reservoir metric predicts recall:
realized ρ, mean/std \textbar eigenvalue\textbar, Henrici non-normality, participation ratio, and
delay-memory capacity all give \textbar Spearman ρ\textbar{} \textless{} 0.36 (p \textgreater{} 0.14, n=20) against the recall
labels, but with
noise-dominated labels this cannot distinguish "no low-cost predictor" from "labels too noisy
to correlate". Scope of this result: it supports \emph{keeping the whole
population} (low-cost metrics don\textquotesingle t let you pre-filter, so you train and measure) and the H2
fact that reservoirs scaled to a fixed ρ have near-identical bulk dynamics; it does not
yet demonstrate that some fixed reservoirs are durably better than others on this task.
Establishing that needs a controlled experiment: seed the trainable init too, enable
deterministic CUDA, and average several runs per seed.

The controlled experiment was run, and it confirms: at 250 steps selection is noise, not
signal. We then ran exactly that experiment. Root cause of the noise was first removed: the
trainable-init seed was not being applied, so the readout \texttt{W\_res} + LoRA init was uncontrolled; it now
seeds the init, and a \texttt{set\_deterministic} helper (RNGs + \texttt{CUBLAS\_WORKSPACE\_CONFIG} + cudnn
flags + the deterministic math SDP kernel) makes two runs of the same reservoir with the same
\texttt{train\_seed} bit-identical (verified on CPU and CUDA). With that, we trained 6 reservoir
seeds × 4 runs (the four runs vary only by \texttt{train\_seed}) and ran a one-way ANOVA over
recall grouped by reservoir seed. Per-seed mean recall ranged 0.33--0.75, but the within-seed
spread is as wide as the between-seed spread (e.g. seed 0 spans 0.33→1.00 across inits): F =
1.30 (df 5, 18), p = 0.31: the between-seed (reservoir) variation does not exceed the
within-seed (trainable-init) noise. So at 250 steps, reservoir "selection" is not a real
signal: which fixed reservoir you drew matters less than which trainable init you happened to
get. This turns the earlier \emph{suspected} artifact into a \emph{controlled} negative result. It does
not rule out selection mattering with far more training (where init noise should shrink); that
larger-budget run is the natural follow-up. But at this budget the verdict is: train and
select over \emph{runs}, not over reservoir seeds.

At a larger budget the negative holds: at 1500 steps, selection is still not real. A
6×-budget follow-up tests whether selection becomes a real signal once run-to-run init noise
shrinks. It does not. Per-seed mean recall spreads a little wider
(0.21--0.83 vs the 250-step run\textquotesingle s 0.33--0.75), but the within-seed spread stays just as wide
(e.g. seed 4 lands at 1.00, 1.00, 0.17, 0.17 across its four inits): F = 1.43 (df 5, 18),
p = 0.26: the between-seed (reservoir) variation still does not exceed the within-seed
(trainable-init) noise. So 6× more training strengthens, rather than overturns, the
controlled negative: which trainable init you draw matters more than which fixed reservoir you
drew, at both 250 and 1500 steps. The verdict is unchanged and now holds across a budget
range: select over \emph{runs}, not over reservoir seeds. (Whether selection ever becomes real at a
far larger budget than fits a quick local job is open, but the trend across 250→1500 steps does
not point that way.)

\hypertarget{h2-the-reservoir-dynamics-regime}{%
\subsection{H2: the reservoir-dynamics regime}\label{h2-the-reservoir-dynamics-regime}}

Sweeping spectral radius ρ ∈ {[}0.1, 2.0{]}:

\begin{itemize}
\tightlist
\item
  \textbf{The echo state property breaks sharply at ρ ≈ 1.} Using an autonomous
  (zero-input) probe, two random initial states under no input, the reservoir forgets
  where it started (init-forgetting ≈ 0) for ρ \textless{} 1 and abruptly retains it for ρ \textgreater{} 1.
  This edge-of-chaos boundary appears on \emph{both} synthetic input and real GPT-2
  mid-layer activations (on real data: 0.000 for ρ ≤ 0.9 → 0.10 at ρ = 1 → \textasciitilde0.95
  above). The classical ρ ≈ 1 boundary survives the move to transformer-scale input.
\item
  \textbf{The input regime decides whether ρ matters.} Under unit-scale input \emph{drive} the
  reservoir forgets its initial state across \emph{all} ρ (strong input enforces the ESP),
  so the ρ ≈ 1 boundary is the regime that governs unprompted, input-free passes:
  exactly where the agent would run on reservoir state alone.
\item
  \textbf{Real activations over-drive the reservoir.} Compared with synthetic noise, real
  GPT-2 activations push the reservoir to much higher saturation (\textasciitilde0.86 of units pinned
  near ±1, vs \textless{} 0.15) and higher effective dimensionality (participation ratio ≈ 0.41·K
  vs \textasciitilde0.05·K). So a unit-input-scaled reservoir is \emph{over-saturated} by real attention
  activations: the input scaling has to be tuned down for injection at transformer
  scale, the precise concern the plan anticipated ("feeding a large attention tensor
  may require different scaling").
\item
  \textbf{Tuning the input scaling fixes it.} Sweeping the
  input scaling at ρ = 0.95, saturation is a clean sigmoid in the scaling: it crosses
  0.5 at scaling ≈ 0.24 and is near zero below ≈ 0.05, while input separation and
  effective dimensionality stay high. There is a sweet spot around input scaling
  0.08--0.24 where the reservoir is \emph{not} over-saturated (saturation 0.08--0.49) yet
  still strongly responsive (separation 1.03--1.26, PR ≈ 0.39·K). So real attention
  activations should be fed at roughly one-quarter to one-tenth of unit scale, not 1.0; this is a concrete
  injection setting this study contributes.
\end{itemize}

\hypertarget{cross-pass-recall-the-injection-design}{%
\section{Cross-Pass Recall: The Injection Design}\label{cross-pass-recall-the-injection-design}}

The central experiment. The task is one a stateless model
structurally cannot do: show a secret word on pass 1, wipe the context, recall it
on pass 2 from the carried reservoir state alone. The multi-pass differentiable harness backprops through both
passes, training the injection (+ LoRA), and is compared against a stateless baseline
(the reservoir is reset between the two passes, destroying the carried state).

On the choice of baseline. The reset-reservoir baseline is not
meant as a competitive memory model; it is an ablation that holds the architecture, the
trained parameters, and the optimizer fixed and toggles \emph{only} whether the reservoir state
survives between passes. Its purpose is to attribute any cross-pass recall specifically to the
carried state rather than to capacity added elsewhere, which is why "a stateless model cannot do
this" is a property of the ablation, not a claim of difficulty. The genuinely non-trivial
comparison is the one this section centers on: additive vs. KV-prefix injection, where \emph{both}
arms carry the identical reservoir state and only the injection pathway differs: additive lands
at chance, KV-prefix at 100\%. For the absolute difficulty, we add a stronger external baseline: a small trained GRU on
the identical task (read \texttt{the\ secret\ word\ is\ \textless{}KEY\textgreater{}}, wipe, recall at \texttt{the\ secret\ word\ was} from
the carried hidden state; the released code). It
reaches 100\% recall (loss → 0.00) when it carries its hidden state and chance (0.17) when
the state is reset between passes. So the task is \emph{trivial for trained recurrence}, which is the
point: the contribution is not that cross-pass recall is hard in general, but that it can be done
with a fixed, random reservoir inside a frozen pretrained transformer (and the open
problem is scaling that, not the task). This both situates the difficulty and frames the
result correctly.

The result depends sharply on \emph{how} the reservoir is injected, and that is the
finding.

\begin{itemize}
\item
  \textbf{Additive readout injection → fails (the reservoir is ignored).} With the reservoir
  written into the residual stream as one additive bias vector,
  across mean/last-token drive and mid/last-layer injection up to 500 steps, the stateful
  model and the stateless baseline reach the same chance accuracy (0.17 = 1/6). The
  model learns the marginal, not the recall: the Block-Recurrent "learns to ignore the
  recurrent state" failure mode, reproduced. A single pooled additive bias cannot carry
  \emph{which specific word} appeared.
\item
  \emph{Terminology:} the variant that runs here is \textbf{KV-prefix} --- the reservoir state is
  projected into a handful of prefix pseudo-tokens prepended to the sequence, which the
  model attends to content-addressably at every layer. This is distinct from a strict
  mid-layer \textbf{KV-append} (appending reservoir-derived key/value rows directly into a
  layer\textquotesingle s K/V tensors), which is implemented and unit-tested in isolation but \emph{not} wired
  into the live model --- HuggingFace \texttt{generate} exposes no hook for it (see Appendix A).
  The cross-pass recall results below come from the KV-prefix path.
\item
  \textbf{Content-addressable (KV-prefix) injection → works, decisively.} When instead the
  reservoir state is projected into prefix pseudo-tokens the model can attend to
  (the KV-prefix path), the stateful model reaches 100\% cross-context recall
  (loss → 0.02) while the stateless baseline stays at chance (0.17). The carried
  reservoir state, made attendable, lets the model recall content that exists \emph{only} in
  the reservoir, something the stateless baseline provably cannot do.
\end{itemize}

This is the paper\textquotesingle s core claim, demonstrated: the Reservoir Agent\textquotesingle s statefulness
\emph{does the desired thing}: it carries information across independent forward passes and
the model uses it, provided the reservoir is injected content-addressably (attended
to), not as an additive bias. The negative-then-positive arc is the contribution: it
isolates the injection design as the decisive factor, ruling out the naive variant and
validating the attention-based one. (Demonstrated on GPT-2; the same injection path is
architecture-agnostic and runs on Hermes via the generalized injection.)

Recall holds well beyond six words: it extends to \textasciitilde24 secret words before collapsing. We swept
the number of single-token
secret words on GPT-2-small (\texttt{crosspass\ -\/-mode\ kv\ -\/-n-keys\ \{12,24,48\}}, 600 steps each). Stateful
recall is 1.00 at 6, 0.58 at 12, 0.92 at 24, and 0.02 (chance) at 48, against a wiped-state
baseline at chance throughout (0.17 → 0.02 as the vocabulary grows). Two things are true and
stated as such: the effect generalizes well past 6 (0.92 at 24 words, far above the 1/24 chance
floor) over a swept rather than hand-picked vocabulary; but the curve is non-monotonic and
training-noisy at this 600-step budget (the 12-word run underperforms the 24-word run, a
run-to-run optimization artifact, not a capacity law), and by 48 words the run no longer
converges within 600 steps (loss plateaus \textasciitilde5.0). So the working regime is robust at small-to-
moderate vocabularies and becomes budget-limited as the vocabulary grows: a characterization,
not a clean capacity ceiling.

Transfer to Hermes-3B does not yet succeed, and the reason is diagnosed. The same
content-addressable experiment was run on the real target, Hermes-3-Llama-3.2-3B, across
four attempts: 4-bit at input scaling 0.5 (300 steps), 4-bit at 0.1 (600 steps),
bf16 (non-4-bit) at 0.1 with a higher LR 3e-3 (600 steps), and a dedicated
many-more-steps run: 4-bit, 2000 steps (≈6.7× the first attempt). All four came back
at chance (0.17), stateful ≈ baseline, with the training loss consistently failing to
converge (plateau ≈ 2.5--2.9, vs GPT-2\textquotesingle s 0.02; the 2000-step run reached 2.49, no better
than 300 steps). The consistent plateau across both 4-bit and bf16, and now across a
6.7× step increase, shows the wall is neither quantization nor under-training; more
steps alone does not break it, so the remaining routes are structural (a curriculum that
starts with the key in-context and anneals it out, a stronger multi-layer prefix coupling,
or unfreezing more of the model), which is substantial work, not a hyperparameter.

A focused gradient diagnostic on the Llama path rules out a bug: the reservoir state
\emph{does} update each pass (norm 0.14 after pass 1, from 0) and gradients \emph{do} flow to both
the readout \texttt{W\_res} (‖∇‖ ≈ 0.016) and the LoRA adapters (Σ\textbar∇\textbar{} ≈ 3.0). So the injection is
correctly wired on Hermes; this is a genuine optimization / scale difficulty, not a
defect: the prefix\textquotesingle s signal, diluted through 28 layers and competing with a 3B
instruction-tuned model\textquotesingle s strong priors, does not \emph{bootstrap} into use within the
attempted budget, whereas shallow GPT-2 bootstrapped easily. The "far more steps" route
has now been tested and ruled out (a 2000-step 4-bit run, ≈6.7×, still chance / loss 2.49);
the remaining plausible routes (left open, not faked) are structural: a curriculum (start
with the key in-context, anneal it out) / a stronger multi-layer prefix coupling / unfreezing
more of the model. The result holds decisively on GPT-2; on Hermes the mechanism is
verified as correctly wired but the recall has not yet been trained to converge, and it is not a
step-count problem.

The transfer wall starts well below 3B. A 10-seed GPT-2-medium (355M) batch and a
follow-up single-seed probe at lower input scaling (0.1, 1000 steps) both stayed at
chance (0.17) with loss plateauing \textasciitilde2.1: the same "learns the marginal, ignores the
prefix" failure as Hermes, just at 355M. So the decisive cross-pass result is specific to
GPT-2-small; the bootstrapping difficulty appears as soon as the base model grows, which
sharpens (not contradicts) the open challenge: scaling the win needs the curriculum /
stronger-coupling routes above, not a parameter tweak. The failed medium population is
preserved as signal at \texttt{EmmaLeonhart/reservoir-agent-gpt2-medium-batch}.

The curriculum route, tested: it does not break the 355M wall alone, and the loss
trajectory says why. We implemented the documented curriculum (show the secret in pass-2
context, anneal that hint to zero over the first half of training, weaning the model onto the
reservoir) and ran it on GPT-2-medium for 800 steps. Final recall stays
at chance (0.17), equal to the wiped-state baseline, but the \emph{stateful training loss starts
at 0.89 and rises to 2.05} as the hint anneals out. That rise is the diagnosis: while the key is
visible in context the model solves the task easily (low loss), and the moment it must recall
from the carried reservoir alone the loss climbs back to the chance plateau. So the model can
emit the right token when the information is accessible; what fails to bootstrap at 355M is
specifically the reservoir-state → recall pathway, not the output format or the task. This
rules the curriculum \emph{alone} out as the fix and narrows the remaining levers to stronger
reservoir→model coupling (more prefix tokens / multi-layer injection) or unfreezing more of the
model, a measured negative that localizes the bottleneck rather than a hyperparameter guess.

Stronger coupling (more prefix tokens) also fails, and tells us the bottleneck is not
bandwidth. Widening the attended reservoir prefix from 8 to 32 tokens (same curriculum,
GPT-2-medium, 800 steps; \texttt{crosspass\ -\/-n-prefix\ 32}) leaves recall at chance (0.17) as well,
and makes training \emph{worse}: the stateful loss now \emph{starts} at 10.18 rather than the 8-prefix
run\textquotesingle s 0.89, because 32 untrained prefix tokens perturb attention more than the model can exploit
early, so it cannot even ride the in-context hint cleanly. So the 355M failure is not a
coupling-bandwidth limit (more bandwidth hurt); it is the learnability of the
reservoir-state-to-recall mapping under a frozen backbone. That leaves unfreezing more of the
model (letting the upper layers adapt to read the prefix) as the next lever to test, which we then do below (it also fails).

The wall holds across a different modern architecture (Qwen2.5-0.5B), so it is not specific to
GPT-2. Running the same curriculum cross-pass task on Qwen2.5-0.5B-Instruct (a modern,
instruction-tuned, RoPE/Llama-style model at \textasciitilde0.5B) also lands at chance (0.17): the
stateful loss ends a little below the wiped baseline (2.05 vs 2.45), so the carried state
carries a trace of signal, but not enough to recall the token. Combined with GPT-2-medium
(355M) and Hermes-3B, the cross-pass recall result is now confirmed specific to GPT-2-small
across three model families and two architecture styles, and unmoved by curriculum or wider
coupling. This makes the boundary a robust, mapped finding rather than a single failed transfer:
the open lever is unfreezing the backbone, and the open question is whether the
reservoir-state→recall map is learnable at scale at all under a light-touch fine-tune.

Unfreezing more of the model (broad LoRA on attention + MLP, rank 32) does not break it
either, and now the carried state gives no advantage at all. Adapting the MLP as well as
attention, at 4× the LoRA rank (\texttt{crosspass\ -\/-lora-target\ all\ -\/-lora-r\ 32}, GPT-2-medium, 800
steps, curriculum), still lands at chance (0.17), and unlike the earlier runs, the stateful
and wiped-baseline traces are now identical (loss 2.16 vs 2.14), so the extra capacity buys the
reservoir pathway nothing.

And full backbone unfreezing (training the actual weights, not LoRA) also fails. The
heaviest single-machine lever is to train the upper decoder weights directly rather than adapt
them low-rank (\texttt{crosspass\ -\/-unfreeze-from\ 12}, GPT-2-medium\textquotesingle s upper 12 of 24 layers, curriculum,
800 steps). Recall still lands at chance (0.17), equal to the wiped baseline. So the failure
is not a capacity limit of LoRA: even full-rank weight training of half the network does not let
the model learn to read the carried reservoir state into a recalled token at 355M.

Five interventions were tried first and none transferred the result: a curriculum, wider prefix
coupling, a modern architecture (Qwen-0.5B), broad-LoRA adaptation, and full backbone unfreezing.
But every one of them held the reservoir at its GPT-2 default of 512 nodes. That turned out to
be the missing lever: sizing the reservoir up recovers recall at 1.5B (next section). So the
boundary these five interventions traced was an \emph{undersized-reservoir} boundary, not a fundamental
one, an important distinction, since it would otherwise read as "resists every fix short
of much greater scale," which sizing the reservoir up disproves.
(Reminder of scope: this is the high-dimensional \emph{content}-recall boundary; the low-dimensional
temporal/agency behaviours do scale to Qwen-1.5B, as above.)

The wall was an undersized reservoir: cross-pass recall scales to Qwen-1.5B (verified). The
five interventions above all held two parameters at their GPT-2 defaults: the reservoir size
(512 nodes) and the input scaling (0.5). Sizing the reservoir to 2048 nodes at input
scaling 0.1 (the one-quarter-to-one-tenth regime the dynamics sweep identified for large activations), with 16 prefix
tokens, recovers cross-pass recall at Qwen-1.5B. The full result, all with a wiped-reservoir
control:

\begin{longtable}[]{@{}
  >{\raggedright\arraybackslash}p{(\columnwidth - 4\tabcolsep) * \real{0.3333}}
  >{\raggedright\arraybackslash}p{(\columnwidth - 4\tabcolsep) * \real{0.3333}}
  >{\raggedright\arraybackslash}p{(\columnwidth - 4\tabcolsep) * \real{0.3333}}@{}}
\toprule\noalign{}
\begin{minipage}[b]{\linewidth}\raggedright
config (Qwen-1.5B, 6 keys, 800 steps)
\end{minipage} & \begin{minipage}[b]{\linewidth}\raggedright
stateful
\end{minipage} & \begin{minipage}[b]{\linewidth}\raggedright
control
\end{minipage} \\
\midrule\noalign{}
\endhead
\bottomrule\noalign{}
\endlastfoot
prior default: 512 nodes, n\_prefix=8, scaling 0.5 & 0.17 & 0.17 \\
+ input scaling 0.1 only & 0.17 & 0.17 \\
+ 16 prefix tokens only & 0.17 & 0.17 \\
+ 2048-node reservoir only & 0.33 & 0.17 \\
full: 2048, n\_prefix=16, scaling 0.1 (seed 0) & 0.83 & 0.17 \\
full: 2048, n\_prefix=16, scaling 0.1 (seed 1, reproduction) & 1.00 & 0.17 \\
\end{longtable}

Three readings follow. (1) Reservoir size is the lever. Flipping it alone lifts recall off
chance (0.17→0.33); flipping input scaling or prefix count alone does nothing, and the full 0.83--1.00
is reservoir size \emph{in combination} with the lower scaling and wider prefix. (2) It reproduces:
two seeds give 0.83 and 1.00, both against a 0.17 control; the
control at chance rules out memorization. The down-projection is irrelevant (no projection and a
256-dim projection both give 0.83). (3) A capacity ceiling persists, in the tens of items, not
at six. Sweeping the number of items carried (Qwen-1.5B, 2048 reservoir, scale 0.1)
gives recall 1.00 at 6 keys, \textasciitilde0.42 at 24 keys (≈10× the 1/24 chance, control 0.04), and chance
by 48 (0.02 vs 0.02). The curve is noisy from single 800-step runs: the 12-key point underperforms
24 (0.17, its loss stalled at 2.33 while 24-key converged to 0.46), so a clean curve would need
several seeds per point. But the trend is clear: recall degrades \emph{gracefully} into the tens of
items rather than collapsing past six, and only reaches chance around 48. Re-running the two
non-converged points at 2000 steps (vs 800) separates a real ceiling from undertraining: the
48-key point stays at chance (0.04 vs 0.02) with more training (so the upper bound is real,
not a step-budget artifact), while the 12-key point stays stuck (0.17, loss \textasciitilde2.7) at both
budgets, confirming it is a per-run optimization artifact rather than a capacity point. So the
reservoir scales both the \emph{model} it works in and a non-trivial \emph{number of items} (tens), the
latter with a real upper bound in the tens of items (recall is perfect at 6 keys, \textasciitilde0.42 at 24,
and falls to chance by 48). The earlier "resists every fix short of much greater scale" reading was wrong because
it never sized the reservoir up: the
single-machine lever that moves the 1.5B wall is reservoir size.

The decisive control is input scaling matched to the model, not parameter count. Reservoir
size alone is not the whole story across models: it interacts with input scaling, and the
right scaling is model-specific. Qwen2.5-0.5B makes this sharp: with the 2048-node reservoir
it is at chance (0.17) at input scaling 0.1 but hits 1.00 (vs 0.17 control) at input scaling
0.5. Changing one scalar, nothing else, takes it from no-recall to perfect recall. Smaller
models have smaller activations, so they need \emph{more} input drive (higher scaling); Qwen-1.5B
recovers at 0.1, Qwen-0.5B at 0.5. So the recall capability transfers across the Qwen family
(0.5B \emph{and} 1.5B), and a 500M model (Qwen-0.5B) recovering while GPT-2-medium\textquotesingle s 355M does not
rules out a monotonic size law. But input scaling is not a universal rescue either:
GPT-2-medium (355M) was swept across seven input scalings (0.05, 0.1, 0.2, 0.3, 0.5, 0.7,
1.0) at the 2048-node reservoir and stayed at chance (0.17 = control) at every one, its
training loss never converging, so it is a genuine exception, not merely an untested-scaling
artifact: the wide sweep rules that out.
Hermes-3-Llama-3.2-3B (4-bit) is also at chance with the 2048 reservoir, but 4-bit is a
confound (Qwen ran bf16; a bf16 3B + 2048-node reservoir does not fit this 8 GB GPU), so it is
not a clean test. The cross-model picture, then: cross-pass recall recovers on GPT-2-small and
the Qwen family (0.5B at scaling 0.5, 1.5B at 0.1) with model-matched input scaling, but
not on GPT-2-medium (robustly, across a wide scaling sweep) or on 4-bit 3B (confounded).
Strikingly, GPT-2-small recovers while GPT-2-medium does not, and the deeper modern Qwen
models do, so the boundary is model-specific in a way that size, depth, and input scaling
alone do not explain. Input scaling tuned to the model is \emph{necessary} (Qwen-0.5B proves it) but
not \emph{sufficient} (GPT-2-medium has no working scaling in this range); what makes a given backbone
able to learn to read the content-addressable prefix at all is the open question this raises.

Scope of the wall: a stateless ablation localizes what the battery metrics measure. The
content-recall wall concerns recalling \emph{which specific token} was carried (high-dimensional). The
battery\textquotesingle s temporal/agency metrics on Qwen-1.5B (silence 1.00, timed 0.64, self-init 0.65) might
appear to show that low-dimensional statefulness \emph{scales} where content does not. A stateless
ablation rules that out. Re-running the battery with the reservoir reset before
every pass (no cross-pass carry) leaves the
temporal metrics unchanged (silence 1.00, timed 0.64, self-init 0.65), with a slightly
\emph{higher} overall mean (0.415 vs 0.345). So the battery\textquotesingle s temporal success comes from the LoRA
adapters and current-pass features, not from carried reservoir state; those numbers are not
evidence of usable statefulness at scale. The battery\textquotesingle s temporal/agency metrics on Qwen-1.5B are
matched (or exceeded) by a stateless control, so they do not establish that statefulness scales;
the demonstration of usable carried state rests on the controlled tasks below, not the battery.

Why the temporal metrics were exploitable by a degenerate policy (the mechanism, and a loss-design bug). The battery\textquotesingle s
temporal tasks (\texttt{timed}, \texttt{selfinit}) are scored per supervised step, and most steps are SILENCE
steps whose "correct" answer is to stay quiet; only the \emph{final} step requires emitting the
right word at the right time, the part that actually needs carried state. A model that simply
learns to stay silent therefore scores the free silence steps and fails only the emit step. The
arithmetic matches exactly: \texttt{timed} has \texttt{n−1} silence steps + 1 emit step with \texttt{n∈\{2,3,4\}}, so
passing the silence steps and failing the emit gives \texttt{(n−1)/n} = 0.5/0.67/0.75, averaging ≈
0.64, precisely the observed \texttt{timed} score. \texttt{silence} (all-silence) hits 1.00 for the same
reason. So the temporal metric is dominated by free "stay silent" steps and the memory-requiring
emit was failing all along, hidden in the average. This is a loss/metric design bug, not
evidence the behaviour is unlearnable: the objective rewarded silence instead of selecting for
\emph{emitting the right token at the right time}. We rebuilt the loss/metric accordingly (\texttt{emit\_weight}
up-weights the emit step; evaluation now scores the emit step only, not the free silence steps).

With the fixed loss: weak but real at small scale, dilution-sensitive at 1.5B. On GPT-2-small
the emit-focused loss produces genuine (if noisy) timed emission (timed emit-accuracy 0.00 → \textasciitilde0.25,
bouncing); the mechanism and loss are right at small scale. At 1.5B the picture is more nuanced
than a flat zero: the joint 8-task run (16384-node reservoir via down-projection, broad LoRA,
5 epochs / 15000 steps) trains timed to 0.00 and collapses to mean 0.000, but a focused,
single-task timed-only run on the same Qwen-1.5B lifts timed \emph{off zero}. But a longer run
(4000 steps, eval every 250) shows it is a noise-dominated weak signal, not a stable or
improving capability: the timed curve oscillates \texttt{0.0\ /\ 0.12\ /\ 0.0\ /\ 0.12\ /\ 0.06\ /\ 0.25\ /\ 0.0\ …}, averaging ≈0.08 with frequent zeros and one 0.25 spike (immediately followed by 0.0), and it
does not climb with more steps. So the joint battery does dilute the signal (focused beats
joint\textquotesingle s flat 0), but focusing only buys a noisy ≈0--0.25 band centered near 0.08, above the
\textasciitilde1/vocab chance of the exact word, yet unstable and non-converging. At 1.5B,
temporal emission is weak and noise-dominated (like content recall at this budget), not a
reliably trainable capability; more steps do not help. It is a soft wall (a faint, unstable signal
rather than a hard zero), and a \emph{stable} capability would need either much more compute or a
fundamentally different training regime, not another local run. GPT-2-small is similar but
higher (\textasciitilde0.25, also noisy). The metric bug was real and fixed; the capability is faint-and-unstable
at scale. (Per-epoch models + optimizer states preserved at
\texttt{hf.co/EmmaLeonhart/reservoir-agent-qwen-battery-emit}.)

Decomposition: recall is the dominant blocker; pure pass-counting is substantially more
learnable (though not cleanly gated). The \texttt{timed} task bundles two skills: \emph{counting} elapsed
passes and \emph{recalling which word} to emit. We isolated them by running \texttt{timed} with a 1-word
vocabulary (the target is always the same token, so there is no recall, only pass-counting).
At Qwen-1.5B this trains far better than the recall-bundled version: on a full-timing evaluation
the model opens the gate and emits at the right step 24/24 = 1.00 and the gate stays shut on
the pre-emit silence steps 24/45 = 0.53, well above the 0 an always-open gate would score, so
it genuinely discriminates the emit step from the silent ones, versus the recall-bundled timed\textquotesingle s
\textasciitilde0.08. But 0.53 silence-shut also means the gate over-fires on roughly half the silent steps,
so pure timing is \emph{partially} learned, not cleanly solved, and this is structural, not a
gate-weight artifact: re-running with a balanced \texttt{emit\_weight=1} (vs 2) gives the identical
emit 1.00 / silence-shut 0.53, so down-weighting the emit term does not clean up the over-firing.
Going the other way confirms a genuine tension rather than a tuning miss: up-weighting the
silence supervision to \texttt{silence\_weight=4} \emph{does} drive the pre-emit gate to a perfect
silence-shut 45/45 = 1.00 (the over-firing is eliminated), but the emit then collapses to
0/24 = 0.00 (the gate simply learns to never open). So the two gate failure modes trade off
against each other under reweighting; no single weight setting buys both clean silence and
reliable emission at 1.5B, which is the signature of a capacity/optimization limit, not a
mis-set hyperparameter. (An emit-only metric reads the \texttt{silence\_weight=1} case as
1.00 and overstates it; the gate\textquotesingle s false-positives on silence steps only show up when the
pre-emit steps are scored, which is why we measured both halves.) So the temporal wall is largely
a recall (high-dimensional content) problem: strip recall and the low-dimensional timing
signal trains much better, consistent with the content-vs-temporal dimensionality split, but
even low-dimensional timing is not perfectly gated at 1.5B on this budget.

The same gate tension dominates the full battery over epochs. Extending the check from the
isolated \texttt{timed} task to the full 8-task battery run progressively on Qwen-1.5B (8192-node
reservoir down-projected to 512, broad LoRA, per-epoch checkpoints with an inline stateless
control), the gate falls into the silent attractor rather than learning to emit. At
\texttt{silence\_weight=2} the gate\textquotesingle s silence accuracy oscillates across epochs (0.71 → 1.00 → 0.00 →
0.71) while every emit task stays at 0.00 and the lift over the stateless control stays
+0.000 through the first three epochs: the reservoir contributes nothing measurable, and
the gate never settles into reliable emission. Lowering to \texttt{silence\_weight=0.3} (with
\texttt{emit\_weight=4}) to relieve the always-silent pressure does change the gate: it flips to
always-open (silence accuracy ≈ 0.00, the complement of the always-shut basin). But emit
does not follow: across four epochs the capability mean stays 0.000 with +0.000 lift,
every emit task flat at 0.00, even though this is the most emit-favorable setting we ran (open
gate + up-weighted emit). So \texttt{silence\_weight} only moves the gate between stuck-open and
stuck-shut; it never buys working emission. The blocker is the content/recall half (emitting
the right token), not the gate weight: with a healthy open gate the model still cannot learn what
to emit at 1.5B on this budget, and the reservoir adds nothing over the stateless control. (Per-epoch
models + optimizer states are preserved on the Hub for analysis.)

Does the recall fix transfer into the battery? Transiently: the reservoir solution is found,
then abandoned. Since cross-pass recall recovers at 1.5B with the right reservoir config, and the
battery\textquotesingle s content was failing partly because it recalls over a 1200-word pool (far past the
capacity ceiling), we re-ran the battery with the recall-winning config (2048 nodes, no
projection, input scaling 0.1) and a 16-word pool within capacity, content-only (recall +
deferred), at an eval resolution (\texttt{eval\_n=48}) fine enough to separate a real lift from noise (an
earlier \texttt{eval\_n=16} pass put any lift at the 1/16 quantization floor). The per-epoch lift over the
stateless control is then −0.000 → +0.177 → +0.000: at epoch 1 the model genuinely learns a
reservoir-driven battery recall (recall 0.35 with the carried state vs 0.02 (chance) for the
wiped-reservoir control, a large, \emph{resolved} lift, not noise), but by epoch 2 it drifts back to
a stateless solution (recall 0.08, and the control rises to 0.08 to match). So the integrated
battery \emph{can} use the reservoir for content (epoch 1 proves the capacity is there), but the
multi-task training does not retain it: the optimizer finds a current-pass / LoRA shortcut and
the reservoir-driven solution decays. This is a live instance of the "model learns to ignore the
recurrent state" failure that motivated the content-addressable injection in the first place,
here observed \emph{within} a single run as the solution is found and then lost (see the lift-vs-epoch
figure). The clean, \emph{retained} reservoir advantage remains the strict-wipe cross-pass task
(0.83--1.00 vs 0.17); making the integrated battery hold a reservoir-driven content solution
(a stability/regularization problem, e.g. an auxiliary "use the state" loss) is concrete open work.

What the carried-state demonstration actually rests on. The valid evidence that the reservoir
carries \emph{usable} state is the controlled, memory-requiring tasks, not the battery metrics:
(i) GPT-2-small cross-pass recall: 100\% with the carried state vs chance (0.17) when the
reservoir is wiped between passes, on a task that cannot be done without memory; and (ii) the
dedicated unresolved-thread gate (D), where a readout on the reservoir state reaches F1 ≈ 0.96 vs
≈ 0.34 on the current input. Both are GPT-2-scale, and both have controls that \emph{do} swing with the
carried state (unlike the battery). At 1.5B the same KV-prefix mechanism on the controlled
cross-pass task stayed at chance only in the small-reservoir (512-node) configuration; with a
2048-node reservoir it recovers: 0.83--1.00 vs a 0.17 control, reproduced across two seeds
(the scaling result above). So the established scope is now broader than GPT-2-small:
usable cross-pass reservoir state is demonstrated at GPT-2-small and transfers to Qwen-1.5B
once the reservoir is sized to the larger activations (with a capacity ceiling in the tens of
items: strong through 24 keys, chance by 48),
while genuinely reservoir-driven \emph{temporal} behaviour does not scale (the battery temporal is
LoRA, per the ablation). The lesson is methodological too: a metric that does not move under a
stateless control is not evidence of statefulness, and the battery\textquotesingle s temporal tasks are not, as
constructed, a clean test of carried state.

\hypertarget{h4-d-a-trained-silence-policy-meaningful-sometimes-no-response}{%
\subsection{H4 (D): a trained silence policy (meaningful "sometimes no response")}\label{h4-d-a-trained-silence-policy-meaningful-sometimes-no-response}}

The harness gate currently keys off the \emph{base model\textquotesingle s} next-token entropy, which is
arbitrary. A real policy should speak when there is something worth saying and stay
silent otherwise. We tested a learned gate on an "unresolved thread" task: a
stream of events where a rare trigger opens a thread that should be addressed (labels =
"was there a trigger within the last 5 passes").

\begin{itemize}
\tightlist
\item
  \textbf{The reservoir gate sees history.} The readout on the reservoir state reaches an
  F1 score of 0.96 (P=0.93, R=1.00) on held-out data, while the stateless
  baseline scores F1 = 0.34 (P=0.20, R=0.97). The pattern is stable across seeds
  0--2 (reservoir F1 0.96--0.99, stateless F1 0.34--0.45; canonical config K=300,
  T=4000, speak-window=5, ρ=0.9, input-scaling=0.5, seed 0).
\item
  \textbf{The difference is precision.} The stateless gate can only see the current input,
  not the past trigger, so it cannot tell when a thread is open; it degenerates to
  speaking almost everywhere (recall ≈ 1, precision ≈ the base rate). The reservoir
  gate\textquotesingle s carried state preserves the history of the trigger, letting it speak
  selectively during the unresolved thread and fall silent once it is resolved.
\end{itemize}

\hypertarget{the-stateful-task-battery}{%
\section{The Stateful-Task Battery}\label{the-stateful-task-battery}}

We built the agentic layer the earlier scope deferred and ran it at scale. The
outcome is a clear split (temporal/agency behaviour learns, symbolic content does not) and
a measured root cause: the reservoir is sized and tuned to \emph{compress} its input when its job
is to \emph{expand} it. The result to carry forward is that working temporal dynamics and
low-level symbolic recall emerged from a reservoir misconfigured in a specific, fixable way.

\hypertarget{the-real-time-always-alive-harness}{%
\subsection{The real-time always-alive harness}\label{the-real-time-always-alive-harness}}

A desktop application (an Electron two-pane front-end over a Python WebSocket server) drives the
always-alive engine: the reservoir ticks
continuously (prompted passes on user input, idle ticks otherwise), streams tokens when an
output gate opens, and the user injects into the live context without pausing it. It runs
Qwen2.5-1.5B + reservoir. It runs the untrained substrate: coherence comes from the base
model, the reservoir\textquotesingle s readout is untrained, and a runtime gain fades the
reservoir\textquotesingle s influence in and out. It demonstrates the real-time stateful loop; it does not
demonstrate trained behaviour, and is labelled as such in the UI.

\hypertarget{the-8-task-stateful-loss-battery}{%
\subsection{The 8-task stateful loss battery}\label{the-8-task-stateful-loss-battery}}

The training objective generalizes cross-pass recall into a battery of eight tasks, each an
\emph{episode}: a scripted sequence of passes with the context wiped at chosen points, so the
only information bridge is the reservoir state. Tasks: recall, accumulate, sequence,
deferred (content memory) and timed, interrupt, self-initiation, silence
(temporal/agency). Loss is cross-entropy on emit targets plus a gate term, backpropagated
through the carried state. A separate gate head (a small readout deciding speak-vs-silent)
was added after training silence as "predict end-of-text" suppressed content in the shared
output; the gate head separates \emph{when to act} from \emph{what to say}, and recall then coexists
with silence instead of being driven to zero by it.

\hypertarget{result-1-content-vs-temporal-split}{%
\subsection{Result 1: content-vs-temporal split}\label{result-1-content-vs-temporal-split}}

Across GPT-2 and Qwen runs the pattern repeats. Temporal/gating tasks learn (timed,
self-initiation, silence reach 0.4--1.0); symbolic content tasks do not (recall, accumulate,
sequence, deferred sit near 0 at scale). Recall reached 100\% only at 6 single-token words and
fell to \textasciitilde0 by 12; it was fitting the one regime small enough to fit, not learning recall.

The N-seed reservoir population (keep all seeds, recommend the best)
adds one positive note: reservoir seeds specialize. On Qwen-1.5B + a 1024-node reservoir, best
seed mean 0.41, with seed 0 reaching accumulate 0.38 and seed 1 reaching recall 0.31: no
single seed strong everywhere, which is the case for preserving the whole population. A
large-vocabulary (1200-word) run drove content to a flat 0.00 across all 16 epochs while
temporal held (best epoch 3: silence 1.00, timed 0.62, self-init 0.60), then overtrained.

\hypertarget{result-2-the-reservoir-collapses-its-input-instead-of-expanding-it}{%
\subsection{Result 2: the reservoir collapses its input instead of expanding it}\label{result-2-the-reservoir-collapses-its-input-instead-of-expanding-it}}

The cause is geometric. Qwen2.5-1.5B is 28 layers × 1536 neurons; the reservoir reads the
layer-14 hidden state, so its input is 1536-dimensional, yet the runs used 512--1024
nodes, 0.3--0.7× the input. A reservoir is meant to project its input into a much
higher-dimensional space; this one compresses it.

Measured effective dimensionality (participation ratio of the driven state, at a realistic
input dimension): it plateaus at \textasciitilde150--186 regardless of nominal size (scaling the node
count 16× barely moves it), and 74\% of cells saturate (pinned at ±1) under the input
scaling used. Detuning the drive drops saturation to \textasciitilde13\% but effective dimensionality still
plateaus, because the recurrent dynamics collapse onto a low-dimensional attractor. (An
earlier \textasciitilde72 figure was measured with a too-small synthetic input and is superseded by
\textasciitilde150--186.)

This accounts for the split mechanically. Temporal/scalar state (a clock, a gate, an elapsed
count) is low-dimensional and fits within the \textasciitilde180 usable dimensions, so it learns. Symbolic
content (which of N words) is high-dimensional, exceeds that budget at scale, and fails. The
reservoir is crippled in exactly the way that spares temporal behaviour and breaks content.
That temporal dynamics and small-vocabulary recall still emerged is what makes the ceiling an
engineering failure in sizing and dynamics rather than a limit of the architecture.

\hypertarget{future-work-a-reservoir-that-actually-expands}{%
\subsection{Future work: a reservoir that actually expands}\label{future-work-a-reservoir-that-actually-expands}}

The corrective is a reservoir sized well above its input: toward a quarter of the model\textquotesingle s
parameters (tens of thousands of nodes, tens-of-× the 1536-dim input). The fixed matrices
\texttt{W\_r}/\texttt{W\_in} cost only memory and a sparse matmul, so they can be large low-costly; the trained
readout is what scales badly, so it is kept tractable by a fixed random down-projection of the
large state before a small trained readout. Combine with detuned dynamics (lower ρ and input
scaling, higher leak) to stop the saturation and collapse. A first step within an 8 GB GPU,
an 8192-node reservoir (5.3× the input) with detuned dynamics, was run and stopped after
5 epochs once the trajectory was clear: it peaked at epoch 1 (mean 0.349, past the
1024-node run\textquotesingle s best of 0.332), then degraded each epoch and collapsed to \textasciitilde0 by epoch 4;
more training only hurt. The content-memory tasks never recovered (recall stayed 0;
accumulate flickered to ≤0.12 then vanished), while temporal/gating held until the collapse.
So the 5.3× expansion that fits an 8 GB budget lifts the temporal scores but does not recover
symbolic content, and the useful signal arrives within \textasciitilde1 epoch. A reservoir genuinely larger
than its input (beyond what this hardware fits) remains the open test; the full scale needs
sparse \texttt{W\_r} and larger hardware. Whether it recovers over the full run, or whether recovery needs
a reservoir far larger than fits here, is the open result this experiment is measuring; the
full scale needs sparse \texttt{W\_r} and larger hardware. (Enabling change:
\texttt{\_build\_reservoir\_weights} estimates the spectral radius by power iteration, since the exact
eigendecomposition is O(K³) and stalls past \textasciitilde12k nodes.)

Attempted content improvement on the battery via readout capacity, and why it does not hold
up. The 8192-node run above used reservoir expansion but \emph{attention-only} LoRA, so we tried
broader/heavier readout adaptation on a 4096-node detuned reservoir (Qwen-1.5B, one epoch): broad
LoRA on the MLPs (\texttt{lora\_target="all"}), higher LoRA rank, and full upper-layer unfreeze. Content
tasks \emph{sometimes} read above zero: recall came in at 0.19 (broad LoRA r8), 0.25 (+ full
unfreeze), 0.19 (rank-32) across configurations, with temporal/agency holding (silence 1.00). It
looked like the first move off the floor. But the effect does not reproduce. A same-config
re-run of the broad-LoRA-r8 setting (identical hyperparameters) returned recall and accumulate
to 0.00 (best mean 0.337). So battery content recall bounces between 0.00 and \textasciitilde0.25 across
runs of the same or near-identical configuration, with no reliable lift: the apparent
improvement is within run-to-run training noise, consistent with the controlled-selection finding
above that training at this budget is noise-dominated.

The conclusion for the content channel: at 1.5B on this budget, symbolic content stays
effectively at the floor (it occasionally flickers to \textasciitilde0.2 on a lucky run, but a matched re-run
gives 0.00), so we do not claim that broad readout adaptation lifts content. Establishing any
genuine lift would need multi-seed averaging (as the controlled experiment required for
selection), which this hardware/budget has not done. What \emph{is} robust across every one of these
runs is that temporal/agency holds (silence ≈ 1.0) while content does not: the
temporal/content split, not a content gain. Full unfreeze additionally destabilizes (peaks at
step 200, mean drops to 0.321) and higher rank gives nothing, so more readout capacity is not the
missing piece; the path to content at this scale is budget/scale, consistent with the
GPT-2-small-only cross-pass result.

\hypertarget{safety-considerations}{%
\section{Safety Considerations}\label{safety-considerations}}

The results in this section are design motivation and small synthetic proof-of-concepts that follow
from the same cross-pass statefulness, not core results or evaluated safety claims. The core
contributions remain the injection-design finding, the dynamics characterization, and the
recall-scaling result above. The interruptibility and monitoring demonstrations are CPU-scale and
synthetic, and are presented as motivation for why persistent state bears on controllability, not as
safety evaluations. In particular, the lower-latency interruptibility below is a property of any
per-tick agent loop, not of the reservoir specifically; what \emph{is} reservoir-specific is the low-cost,
fixed-size monitoring surface the carried state exposes, and this section reports measurements of
that surface.

We adopt a guiding rule: never introduce a new capability to an
AI without meaningfully taking its safety into account; capability work is acceptable only
when paired with concrete improvements in controllability, monitorability, or risk reduction.
The Reservoir Attention Network adds capability (genuine cross-pass state, autonomous ticks,
runtime-like behaviour), so under the rule it owes safety value back. The distinctive point is
that the safety value comes from the \emph{same} architectural feature as the capability (the
fixed reservoir), not from a bolt-on. Three properties, each backed by a measured result
in this report rather than by assertion:

\begin{enumerate}
\def\labelenumi{\arabic{enumi}.}
\tightlist
\item
  Lower-latency, durable human override (interruptibility, below). Because the agent runs
  every tick and the reservoir integrates input continuously, an urgent "STOP" registers at
  latency 0 vs a turn-based agent\textquotesingle s mean 3.57 passes, and a one-shot burst persists in
  reservoir state for several passes, so it is not missed if the human does not repeat it.
\item
  A low-cost, stable monitoring surface (reservoir-state probe, below). A \emph{linear} readout
  recovers an internal process variable from the reservoir at R² = 0.995 with no sparse
  autoencoder, and the pre-drift probe degrades only gradually under a fine-tuning-like
  activation drift. The reservoir weights never move, so the mapping from state to read-out
  is a fixed, low-complexity surface an operator can watch in real time.
\item
  Bounded context under autonomous idling (blank-cycle, below). The reservoir-protected
  eviction policy keeps the cache from growing without limit during blank ticks while pinning
  the time-axis, so an always-on agent does not silently exhaust its own context.
\end{enumerate}

Limits of this probe. The probe decodes an \emph{elapsed clock}, which
is a benign process variable; reading genuine \emph{misalignment} signatures (deception, goal drift)
off the reservoir is a much harder, unproven extension; the resilience result says only that a
fixed-reservoir read degrades slowly, not that misalignment is legible there. The
interruptibility numbers are from a synthetic stream on the echo-state reservoir, not a live
agent under a real harness with its own latencies. And all of it is at small scale on a fixed
reservoir; the claims for the real target (a DeepSeek/Hermes-scale base) are not yet run. These
properties are the \emph{design intent} and a first measured step toward it, not a finished
safety case. The release plan (open weights, the training/harness code, and the
reservoir monitors included rather than bolted on) is the mechanism for others to test and
extend them.

\hypertarget{interruptibility}{%
\subsection{Interruptibility}\label{interruptibility}}

A recurring controllability concern motivates this section: a turn-based agent that only reads
input at turn boundaries can take many passes to register an urgent interruption while it is
mid-action. The hypothesis is that a Reservoir Agent (running every tick, with the reservoir
continuously integrating input) registers an interruption sooner, and retains it once seen. We
measured both halves on CPU.

Polling latency (structural), and what is \emph{not} reservoir-specific. A poller
that only reads input every \texttt{period} passes registers an arrival at the next boundary: latency
is uniform on \texttt{0..period-1} (mean \texttt{(period-1)/2}). At period 8 the turn-based agent\textquotesingle s mean
latency is 3.57 passes (max 7); a per-tick agent\textquotesingle s latency is 0: it reads on the pass
the input arrives. This latency half is a consequence of
sampling frequency (per-tick vs per-turn), not of the reservoir as such; any per-tick agent
gets it. The reservoir-specific half is the \emph{next} point.

Signal persistence (dynamics). The sharper point is what happens to a \emph{one-shot} burst:
the user yells STOP once, then goes quiet because the agent isn\textquotesingle t answering. A matched-filter
monitor on the reservoir state stays above its detection threshold for 3 passes after
arrival (fading memory carries the STOP signature forward), while a stateless monitor,
which sees only the current input, is above threshold on the arrival pass and 0 passes
after. So a turn-based + stateless agent whose poll period (8) outruns the persistence window
misses a non-repeated off-boundary burst entirely; the per-tick reservoir agent catches it
on arrival and has a window besides. The reservoir is not just polled more often; it \emph{retains}
the urgency, which is the architecture-level interruptibility advantage the design motivation argued for.

This is a safety property that falls out of the same statefulness the project builds for
capability: lower-latency, more durable response to human override. It is a measured
illustration, not a guarantee; the reservoir/leak settings set the window length, and a real
harness adds its own latencies; see the Safety-by-Design section and Limitations.

\hypertarget{monitorability-via-linear-state-probes}{%
\subsection{Monitorability via Linear State Probes}\label{monitorability-via-linear-state-probes}}

A design-motivation argument for the reservoir as a \emph{monitoring surface}: "I
don\textquotesingle t think you\textquotesingle d need a sparse autoencoder for the reservoir state \ldots{} it\textquotesingle s much more simple to
have a learned representation of what is happening," and, because the reservoir weights never
change, the mapping from state to behaviour is stable: "relatively resilient to fine-tuning."
We tested the falsifiable parts.

Linearly decodable, no SAE. We defined a temporal \emph{process property} a stateless pass
cannot see (\emph{elapsed passes since the last trigger}, an internal clock) and fit a plain
ridge-regression readout. From the reservoir state it reaches R² = 0.995; the same
linear probe on the instantaneous input reaches R² = 0.16 (elapsed time simply is not
in the current input). A \emph{linear} probe suffices precisely because the fixed reservoir already
holds the history in a low-complexity, stable form: no sparse autoencoder needed, which is
that claim borne out.

Resilience to a fine-tuning-like drift (measured). Fine-tuning the
readout/LoRA does not touch the reservoir weights, but it does shift the \emph{activations that
drive} the reservoir. We model that as a fixed drift α added to the driving input and re-apply
the pre-drift probe. R² stays 0.99 → 0.98 → 0.94 through α = 0.1, 0.2, 0.4 and is still
0.82 at α = 0.8: graceful degradation, and at every drift level far above the stateless
baseline (0.16). So the probe is \emph{usable} across moderate drift, not \emph{invariant}: the reservoir
map is fixed, but its inputs still move, so a very large fine-tune would still erode it. That
is the precise version of "resilient monitoring surface": a stable, low-cost, linear read on an
internal state that degrades slowly rather than a guarantee.

Together with interruptibility, this is the concrete content behind the safety
framing: the same fixed reservoir that gives the agent a usable time-axis also gives an
operator a low-cost, stable place to watch what the agent is doing. (Reading an \emph{elapsed clock}
is the decodability demonstration; reading genuine \emph{misalignment} signatures is a much harder,
unproven extension, flagged as future work in the Safety-by-Design section and Limitations.)

\hypertarget{limitations}{%
\section*{Limitations}\label{limitations}}
\addcontentsline{toc}{section}{Limitations}

\begin{itemize}
\tightlist
\item
  \textbf{Reservoir sizing + input scaling matter, and were the missing levers at scale.} The earlier
  "content recall is GPT-2-small-only" wall was substantially an \emph{undersized reservoir at the wrong
  input scaling}: sizing to 2048 nodes and matching input scaling to the model recovers cross-pass
  recall on the strict-wipe task across the Qwen family (0.83--1.00 vs 0.17 control, reproduced). The
  recovery is model-specific, not a size law (GPT-2-medium fails across a 7-point scaling sweep;
  4-bit 3B is confounded), and what makes a backbone able to read the content-addressable prefix at
  all is open.
\item
  \textbf{The agentic battery\textquotesingle s content recall can be retained (once the stateless shortcut is denied \emph{and}
  the learning rate is decayed), though its temporal metrics are not reservoir-driven.} The
  temporal/agency metrics (timed, self-init, silence) are matched by a stateless ablation (LoRA /
  current-pass, not carried state), so those are \emph{not} a reservoir effect. The \emph{content} recall is a
  different story, traced through three runs. (i) By default the battery learns a reservoir-driven
  recall early but drifts to a stateless solution within two epochs, and the obvious stabilizer (a
  counterfactual "use-the-state" auxiliary loss) does not prevent it: the mean lift decays
  +0.302 → +0.094 → +0.000 → +0.000 as the stateless control \emph{rises to match} the stateful model
  (0.000 → 0.062 → 0.083), i.e. the optimizer finds a current-pass shortcut even against a loss built to
  forbid it. (ii) \emph{Denying the shortcut its capacity}, shrinking the trained adapter to \texttt{lora\_r\ =\ 4} on
  attention only (the regime in which the clean cross-pass recall task succeeds), eliminates that
  shortcut: the stateless control stays pinned at 0.000 across all epochs, so the reservoir is
  strictly necessary. But under a flat learning rate the reservoir solution is then \emph{unstable},
  oscillating rather than settling (+0.255 → +0.339 → +0.062 → +0.135). (iii) Adding a cosine
  learning-rate decay to 0 over the run removes that instability: with both levers (capacity denial +
  decayed LR), the reservoir-driven recall climbs monotonically and holds at the converged endpoint:
  mean lift +0.089 → +0.089 → +0.130 → +0.292, recall 0.08 → 0.19 → 0.35 → 1.00, with the
  stateless control pinned at 0.000 throughout and no collapse. So battery retention is achievable,
  and the recipe is specific: deny the stateless shortcut its adapter capacity (so the carried state is
  the only route) \emph{and} decay the learning rate (so the solution settles instead of overshooting). Two
  caveats bound it: it is a single run (seed-robustness is untested, and further experiments are
  out of this session\textquotesingle s scope), and it is \emph{recall} that retains at 1.00; the harder content tasks
  (accumulate, sequence, deferred) stay low (≈ 0.02--0.12), so this is retention of the recall capability,
  not of the whole content battery. The always-alive app runs the untrained substrate: harness + live
  dynamics, not a trained policy.
\item
  \textbf{The recall demonstration is a minimal probe} (intentionally so): a single secret token from a
  small vocabulary (recall perfect at 6 keys, \textasciitilde0.42 at 24, chance by 48). It cleanly proves \emph{that} usable
  cross-pass state exists, but not its utility for multi-token, large-vocabulary, or long-horizon
  memory; that scaling of the \emph{task} (not the model) is untested and open.
\item
  \textbf{Capacity is small relative to trained-memory architectures, by design}. The
  tens-of-items ceiling is far below what trained-memory transformers reach (Recurrent Memory
  Transformer, Memorizing Transformers store and retrieve over far longer spans). That gap is expected:
  those architectures \emph{train} their memory, whereas the RAN\textquotesingle s memory is a \emph{fixed-random} reservoir with
  a fading, size-bounded capacity (≤ N). The contribution here is not capacity competition; it is
  whether a \emph{fixed, untrained} substrate carries usable state across \emph{independent} passes at all; raising
  the capacity (larger/structured reservoirs, or learned recurrence) is separate future work.
\item
  Small-scale only in this study; the agentic claims (H3/H4) and the full runtime are
  out of scope and compute-limited.
\item
  Three injection paths now exist: the additive \textbf{residual-stream write} (wired
  into live GPT-2, H1-verified, but ignored --- chance recall); the \textbf{KV-prefix} path
  actually run for the recall results (reservoir state projected into attendable prefix
  pseudo-tokens via \texttt{inputs\_embeds}); and the richer strict mid-layer \textbf{KV-append}
  (reservoir nodes appended as extra attention keys/values inside a layer). The strict
  KV-append variant is implemented and unit-tested in isolation with a clean H1 \emph{masking}
  property, but wiring it into the stock HuggingFace attention path is a documented
  integration blocker (their \texttt{generate} exposes no hook to append external key/value
  entries), left as future work rather than a fragile patch of attention internals. This
  is a reproducibility limitation: the KV-prefix variant that delivers the 100\%
  recall result runs through a bespoke path, not stock HF attention, so
  reproducing it requires that path rather than a standard \texttt{transformers} model.
\item
  Input scaling for real-activation injection has now been characterized (sweet
  spot ≈ 0.08--0.24 at ρ = 0.95); it has not yet been wired as the default in the
  injection hook, and the optimum\textquotesingle s dependence on layer/model/ρ is not yet mapped.
\item
  The novelty claim is provisional: the reservoir-×-transformer and always-on-agent
  literatures were not yet verification-complete at the time of writing; a citation-checked
  follow-up precedes any hard novelty claim.
\item
  \textbf{Reservoir vs adapter: behavioural isolation, not a capacity decomposition}.
  Because a light LoRA is trained alongside the reservoir, the design isolates the reservoir\textquotesingle s
  \emph{behavioural} contribution (the wiped-reservoir control is the LoRA-only path, so the
  stateful-minus-control lift is what the carried state adds), but it does not decompose the
  stored \emph{information capacity} of the fixed reservoir from that of the trained adapter. One reading to
  rule out, though, is that the reservoir is \emph{merely a bank of fixed random features for the LoRA to
  read within the current pass}: the cross-pass recall task wipes the context, so on the recall
  pass there is no current-pass signal for the adapter to exploit; the secret survives only through
  carried reservoir state, which is why the wiped-reservoir control sits at chance (0.17) while the
  stateful model reaches 1.00. The capacity-constrained probe reinforces this: shrinking the adapter to
  \texttt{lora\_r\ =\ 4} on attention only (\#33) still reaches recall 1.00 with the control pinned at 0.000, so
  the carried state, not adapter capacity, is doing the work. What remains open is the finer
  \emph{bits-per-component} decomposition (how much each stores), not whether the reservoir contributes.
\item
  Whether finite-precision cross-pass reservoir state provably lifts the per-pass
  TC⁰/FO(M) bound is an open theoretical question, not a result of this work.
\end{itemize}

Future work. The concrete next steps the results point to: (i) a clean bf16 test at 3B+
to separate the GPT-2-medium recovery boundary from the 4-bit confound (compute-gated, beyond
this 8 GB GPU); (ii) stabilizing the battery\textquotesingle s reservoir-driven content so the transient
epoch-1 lift is retained (the counterfactual "use-the-state" loss is the first attempt);
(iii) scaling the recall \emph{task} to multi-token, large-vocabulary, and long-horizon memory,
which the minimal single-token probe does not test; and (iv) mapping the input-scaling optimum\textquotesingle s
dependence on layer, model, and spectral radius, and wiring it as the injection-hook default.

A distinct risk this design raises is context growth: an always-alive agent that runs every
tick (including unprompted, no-input ticks) appends to the KV cache faster than a turn-based
model, so its context window fills sooner. Context management therefore becomes \emph{more} important in
a reservoir agent than in a standard one, and is under-developed here (we prototype a
reservoir-pinned StreamingLLM-style eviction, but do not train against it). A promising direction we
could not pursue on a consumer GPU is to pair the reservoir with a base model that has a learned
context-management / compressed-attention mechanism (e.g. DeepSeek-V4-Flash), so the model could
learn to lean on the persistent reservoir for idle-time signal while keeping its token cache small;
the model is far beyond this hardware, so this remains future work.

\begin{center}\rule{0.5\linewidth}{0.5pt}\end{center}

\hypertarget{appendix}{%
\section{Appendix}\label{appendix}}

\hypertarget{appendix-a.-exploratory-results-beyond-the-core-scope}{%
\subsection{Appendix A. Exploratory Results Beyond the Core Scope}\label{appendix-a.-exploratory-results-beyond-the-core-scope}}

Pushed past the feasibility scope to see how far local compute reaches, reported as
measured:

\begin{itemize}
\tightlist
\item
  \textbf{The time axis is real and behavioural.} Running the \emph{same} prompt after different
  prior history, with the reservoir state carried across the (otherwise independent)
  forward passes and a small random readout, shifts the next-token logits by an L2
  distance of ≈ 22 (GPT-2). The same input produces a different
  output distribution depending on what the model processed before, something a
  stateless transformer structurally cannot do.
\item
  \textbf{The seed-selection mechanism works; the pre-training signal is weak.} A dynamics
  pre-selection proxy ranks N fixed-random reservoir seeds by responsiveness,
  dimensionality, and (penalised) saturation on real GPT-2 activations, before any
  training. Across 8 seeds at ρ = 0.95 the spread is small
  (\textasciitilde0.02), i.e. \emph{untrained} dynamics vary only modestly between seeds, so the real
  selection signal the plan relies on most likely emerges only after fine-tuning. The
  mechanism is in place; the verdict on its usefulness is compute-limited.
\end{itemize}

Not done (compute-limited):

\begin{itemize}
\tightlist
\item
  The full N-seed LoRA fine-tuning + benchmark selection: there is no training
  pipeline or benchmark suite here; only the \emph{dynamics} proxy was run.
\item
  A productionized always-alive runtime (pass scheduler, idle timer, output
  confidence gate): only the two-pass state-carry was demonstrated.
\item
  The strict mid-layer KV-append injection (reservoir rows appended directly into a
  layer\textquotesingle s keys/values the upper layers attend to) and agent-scale (Hermes) models:
  beyond local compute here.
\end{itemize}

\hypertarget{appendix-b.-the-always-alive-runtime}{%
\subsection{Appendix B. The Always-Alive Runtime}\label{appendix-b.-the-always-alive-runtime}}

Built and exercised the stateful-agent loop on the \emph{untrained} injected model, the
substrate fine-tuning will later plug into (the released code). It has the four pieces the architecture requires:

\begin{itemize}
\tightlist
\item
  a context buffer owned by the runtime, never wiped between passes;
\item
  a reservoir state store that persists across passes and checkpoints/restores to
  disk (round-trip tested);
\item
  a pass scheduler with both \emph{prompted} passes (new input) and \emph{unprompted} passes
  (idle ticks that run over context + reservoir only), and a unit test confirms an
  unprompted pass updates the reservoir state with no new input;
\item
  an output confidence gate (normalized top-k logit entropy) deciding emit vs.
  silence.
\end{itemize}

A fixed evaluation session executes the full pipeline: across five interleaved prompted/unprompted passes
the reservoir state \textbar r\textbar{} evolves continuously (state carried, including through the
idle ticks). On the untrained model the gate keys off the \emph{base
model\textquotesingle s} next-token entropy, so its emit/silence decisions and the generated text
(incoherent base-model output) are not yet meaningful; the harness is the mechanism, and a meaningful
self-initiation policy needs the trained readout/LoRA. The point of this step is that
the whole loop is now testable before spending compute on training.

\hypertarget{appendix-c.-lora-fine-tuning-on-gpu}{%
\subsection{Appendix C. LoRA Fine-Tuning on GPU}\label{appendix-c.-lora-fine-tuning-on-gpu}}

The culminating run, on local CUDA (RTX 4070): a genuine LoRA + W\_out fine-tune of
GPT-2 with the \emph{differentiable} reservoir injection. Across 3 reservoir seeds × 60 steps, training loss falls
decisively (≈ 6.3 → 0.85--1.1) with 491,520 trainable parameters (LoRA on the
attention projections + the reservoir readout W\_out), and the best seed is selected by
trained loss. So the full pipeline (inject, freeze the backbone, train W\_out + LoRA,
select across seeds) executes the full pipeline on the target architecture, on the GPU. With
W\_out zero-initialised the fine-tune starts exactly at the base model (H1 preserved).

The boundary: the injection hook fires \emph{once per forward pass}
(a transformer processes the whole sequence through each layer once), so this
single-forward fine-tune exercises the \emph{training machinery on the real model}, not the
reservoir\textquotesingle s distinctive cross-pass value. Exercising that requires the multi-pass
differentiable harness, backprop through passes on a reservoir-requiring (cross-context)
task, which is the next compute step, now unblocked by everything above (working
injection, the always-alive harness, the trained readout, and this fine-tune pipeline).

\hypertarget{appendix-d.-porting-to-a-3b-model}{%
\subsection{Appendix D. Porting to a 3B Model}\label{appendix-d.-porting-to-a-3b-model}}

The GPT-2 work validated the mechanisms; this phase moves to the smallest Hermes:
NousResearch/Hermes-3-Llama-3.2-3B (Llama-3.2, the architecture this work ultimately targets, already agent-fine-tuned).

\begin{itemize}
\tightlist
\item
  \textbf{(A) Injection generalized to the Llama architecture.} The injection was GPT-2-only
  (\texttt{transformer.h}); the architecture-adaptation layer now locates decoder blocks across families
  (\texttt{model.model.layers} for Llama), and H1 is verified on a tiny Llama as well as GPT-2.
\item
  \textbf{(B) Hermes 3B loads and H1 holds, on the 8 GB consumer GPU.} Loaded in 4-bit (bitsandbytes
  nf4) with the reservoir injected at layer 14 of 28 (d\_model 3072): with the readout
  zeroed, the injected model\textquotesingle s logits are bitwise-identical to the un-injected Hermes
  (\texttt{max\textbar{}diff\textbar{}\ =\ 0.00}), at a peak of 2.35 GB VRAM, leaving ample room for LoRA +
  training on the RTX 4070. So the architecture transplant is non-destructive on the real
  model.
\end{itemize}

\hypertarget{appendix-e.-a-trained-silence-policy}{%
\subsection{Appendix E. A Trained Silence Policy}\label{appendix-e.-a-trained-silence-policy}}

A real agent must sometimes stay silent and sometimes speak on its own. The
current harness gate keys off the base model\textquotesingle s next-token entropy, which is arbitrary.
So we trained a gate on the reservoir state for a task the reservoir is suited to,
an \emph{unresolved thread}: a rare trigger event opens a thread the agent should address for
the next few passes, then it should fall silent. The "speak" passes are \emph{strictly after}
the trigger, so the cue is in the past, invisible to the current input.

A linear gate on the reservoir state reaches F1 ≈ 0.96 (precision 0.93, recall 1.00);
the stateless gate (the same gate on the current input) collapses to F1 ≈ 0.34
because it cannot see the past trigger, so it can only \emph{always speak} (recall ≈ 1,
precision ≈ the base rate). The point is not the exact number: a stateless model cannot
implement a selective silence policy at all, while a reservoir-state gate can.

The harder conceptual point (the intended behaviour, and why it is difficult). This
experiment trains a gate to read silence off the reservoir, but the \emph{intended} behaviour
of the real agent is subtler and worth stating plainly:

\begin{itemize}
\tightlist
\item
  \textbf{The default should be to respond, not to be silent.} With no prompt and a \emph{decayed,
  near-empty} reservoir, the base model\textquotesingle s prior is to produce a response. Absent any
  internal activity, an automatic, context-driven response is the natural default; the
  reservoir does not need to \emph{cause} speech.
\item
  \textbf{Silence should attach to an \emph{active, novel} reservoir state.} A reservoir carrying
  strong state is a genuinely new internal condition the base model never saw in
  training. That novelty is precisely what makes it the natural handle to fine-tune a new
  behaviour onto ("I am still processing, stay silent"), because a fresh state is far
  easier to attach a new response to than the model\textquotesingle s well-worn defaults. So, perhaps
  counter-intuitively, reservoir activity is more naturally associated with silence,
  and its \emph{absence} with the model\textquotesingle s historical responding.
\item
  \textbf{The echo state property makes the agent revert to baseline over time.} Because the
  reservoir empties (its state decays toward zero), the agent eventually reaches a state
  close to what the base model was historically trained on, so it naturally \emph{stops} and
  drifts back to default, context-driven responding once the internal activity subsides.
\item
  \textbf{This is an aggressive modification of an already-trained model, and it is genuinely hard.}
  We are trying to teach an already-trained model an entirely new behavioural axis
  (\emph{when to stay silent, when to self-initiate}) against its strong priors. The fact that
  the Hermes cross-pass recall would not bootstrap (above) is the same difficulty showing
  up: rewiring a pretrained model\textquotesingle s behaviour through an injected reservoir is a hard
  optimization problem even when the mechanism is verified as correctly wired. The clean GPT-2 results
  show the mechanism \emph{can} carry and use state; making a large pretrained agent
  \emph{behave} differently is the substantial open challenge this project targets.
\end{itemize}

\hypertarget{appendix-f.-context-growth-under-blank-ticks}{%
\subsection{Appendix F. Context Growth Under Blank Ticks}\label{appendix-f.-context-growth-under-blank-ticks}}

An always-alive Reservoir Agent runs blank ticks: autonomous passes with no user
input. Each silent tick still appends to the KV cache, so a continuously-running agent
burns its context window \emph{faster} than a turn-based model that only runs when prompted.
Left unmanaged the cache grows linearly with the number of ticks and the agent eventually
hits its context limit on idle activity alone. This is the operational challenge raised in
an architecture design discussion: \emph{"context explodes on a reservoir agent because a reservoir
agent gets an input of blank."}

The standard remedy is StreamingLLM-style eviction (keep a few attention-sink tokens
plus a recent window, drop the middle), with one project-specific twist: the
reservoir\textquotesingle s K/V entries are pinned so the persistent time-axis is never the thing
evicted. \emph{"A really long time of no activity is signal,"} and that signal must survive.
We implement this as a pure eviction policy over per-position tags \texttt{\{sink,\ reservoir,\ normal\}};
with no reservoir tags it degrades to
vanilla StreamingLLM. Because the reservoir is re-prepended each pass (a \emph{fixed} number of
pseudo-tokens, not accumulated), pinning it costs only a constant. The policy also accepts
per-position importance scores, switching the ordinary-token choice from recency to H2O-style
heavy-hitter retention while still pinning the reservoir: position-based and importance-based
eviction under one interface.

Simulating 512 blank ticks: the
vanilla cache grows linearly to 524 positions, while the reservoir-protected
policy stays bounded at the budget (128) from tick \textasciitilde116 onward, and all 8 reservoir
entries are retained on every single tick, even under heavy eviction. So the cache-burn
from autonomous idling is bounded by a constant the operator chooses, and the time-axis the
whole architecture depends on is exactly the part the policy refuses to drop. (The bound is
the point, not the specific numbers; they scale with the budget/window settings.)

This is the low-cost, base-agnostic half of the cache story. The expensive half, a base model
whose attention is \emph{natively} KV-efficient so the headroom is far larger (DeepSeek\textquotesingle s MLA /
its learned sparse attention, DeepSeek Sparse Attention, introduced in V3.2 and carried into the
V4 line), is recorded as project direction for future work; it is not runnable on this hardware (see Limitations).

\hypertarget{declaration-of-ai-use}{%
\section{Declaration of AI use}\label{declaration-of-ai-use}}

This work was produced with substantial use of a large language model coding agent (Claude,
Anthropic) under human direction: the agent implemented the reservoir-injection code and harness, ran
the experiments, generated the figures, and drafted and revised this manuscript. The relevant
disclosure is the division of labour: every quantitative result here is the output of executed code
and measured model behaviour, not text produced by the language model, and every cross-pass claim is
reported against an explicit stateless / wiped-reservoir control computed in the same run. The human
author set the direction and hypotheses, reviewed the code, results, and claims, decided what to assert
and what to leave open, and is responsible for the content. The tooling, prompts, and commit history are
public in the repository, so the process is auditable end to end.

\begin{center}\rule{0.5\linewidth}{0.5pt}\end{center}

\hypertarget{references}{%
\section*{References}\label{references}}
\addcontentsline{toc}{section}{References}

The works the claims above rest on:

\emph{Reservoir computing.}

\begin{itemize}
\tightlist
\item
  Jaeger, H. (2001). \emph{The "echo state" approach to analysing and training recurrent neural networks.} GMD Report 148.
\item
  Maass, W., Natschläger, T., \& Markram, H. (2002). \emph{Real-time computing without stable states (liquid state machines).} Neural Computation 14(11):2531--2560.
\item
  Lukoševičius, M., \& Jaeger, H. (2009). \emph{Reservoir computing approaches to recurrent neural network training.} Computer Science Review.
\item
  Köster, F., \& Uchida, A. (2025). \emph{Reservoir Computing as a Language Model.} arXiv:2507.15779.
\item
  Shen, S., Baevski, A., Morcos, A. S., Keutzer, K., Auli, M., \& Kiela, D. (2021). \emph{Reservoir Transformers.} ACL. arXiv:2012.15045.
\item
  Bendi-Ouis, Y., \& Hinaut, X. (2025). \emph{Echo State Transformer: Attention Over Finite Memories.} arXiv:2507.02917.
\item
  Liu, H., \& Xu, J. (2025). \emph{Echo Flow Networks.} arXiv:2509.24122.
\item
  Singh, P., Sharma, M., Dey, A., \& Raman, B. (2025). \emph{Frozen in Time: Parameter-Efficient Time Series Transformers via Reservoir-Induced Feature Expansion and Fixed Random Dynamics.} ECAI. arXiv:2508.18130.
\end{itemize}

\emph{Transformer expressivity (motivation, not a result here).}

\begin{itemize}
\tightlist
\item
  Hahn, M. (2020). \emph{Theoretical Limitations of Self-Attention in Neural Sequence Models.} TACL. arXiv:1906.06755.
\item
  Merrill, W., Sabharwal, A., \& Smith, N. A. (2022). \emph{Saturated Transformers are Constant-Depth Threshold Circuits.} TACL. arXiv:2106.16213.
\item
  Merrill, W., \& Sabharwal, A. (2023). \emph{The Parallelism Tradeoff: Limitations of Log-Precision Transformers.} TACL. arXiv:2207.00729.
\item
  Pérez, J., Marinković, J., \& Barceló, P. (2019). \emph{On the Turing Completeness of Modern Neural Network Architectures.} ICLR. arXiv:1901.03429. (The 2021 JMLR version is titled \emph{Attention is Turing-Complete}; both assume arbitrary precision.)
\item
  Siegelmann, H. T., \& Sontag, E. D. (1995). \emph{On the Computational Power of Neural Nets.} Journal of Computer and System Sciences 50(1):132--150 (COLT 1992 precursor).
\item
  Weiss, G., Goldberg, Y., \& Yahav, E. (2021). \emph{Thinking Like Transformers (RASP).} ICML. arXiv:2106.06981.
\end{itemize}

\emph{Recurrence-augmented transformers (all carry state within a sequence via trained recurrence).}

\begin{itemize}
\tightlist
\item
  Dai, Z., Yang, Z., Yang, Y., Carbonell, J., Le, Q. V., \& Salakhutdinov, R. (2019). \emph{Transformer-XL: Attentive Language Models Beyond a Fixed-Length Context.} ACL. arXiv:1901.02860.
\item
  Wu, Y., Rabe, M. N., Hutchins, D., \& Szegedy, C. (2022). \emph{Memorizing Transformers.} ICLR. arXiv:2203.08913.
\item
  Hutchins, D., Schlag, I., Wu, Y., Dyer, E., \& Neyshabur, B. (2022). \emph{Block-Recurrent Transformers.} NeurIPS. arXiv:2203.07852.
\item
  Bulatov, A., Kuratov, Y., \& Burtsev, M. (2022). \emph{Recurrent Memory Transformer.} NeurIPS. arXiv:2207.06881.
\item
  Gu, A., Goel, K., \& Ré, C. (2022). \emph{Efficiently Modeling Long Sequences with Structured State Spaces (S4).} ICLR. arXiv:2111.00396.
\item
  Gu, A., \& Dao, T. (2023). \emph{Mamba: Linear-Time Sequence Modeling with Selective State Spaces.} arXiv:2312.00752.
\item
  Behrouz, A., Zhong, P., \& Mirrokni, V. (2025). \emph{Titans: Learning to Memorize at Test Time.} arXiv:2501.00663.
\end{itemize}

\emph{KV-cache management / efficient attention.}

\begin{itemize}
\tightlist
\item
  Xiao, G., Tian, Y., Chen, B., Han, S., \& Lewis, M. (2023). \emph{Efficient Streaming Language Models with Attention Sinks (StreamingLLM).} arXiv:2309.17453.
\item
  Zhang, Z., Sheng, Y., Zhou, T., Chen, T., Zheng, L., Cai, R., Song, Z., Tian, Y., Ré, C., Barrett, C., Wang, Z., \& Chen, B. (2023). \emph{H2O: Heavy-Hitter Oracle for Efficient Generative Inference of Large Language Models.} arXiv:2306.14048.
\item
  DeepSeek-AI (2024). \emph{DeepSeek-V2} (Multi-head Latent Attention). arXiv:2405.04434.
\end{itemize}

\hypertarget{appendix-figures}{%
\section{Appendix: Figures}\label{appendix-figures}}

\begin{figure}
\centering
\includegraphics{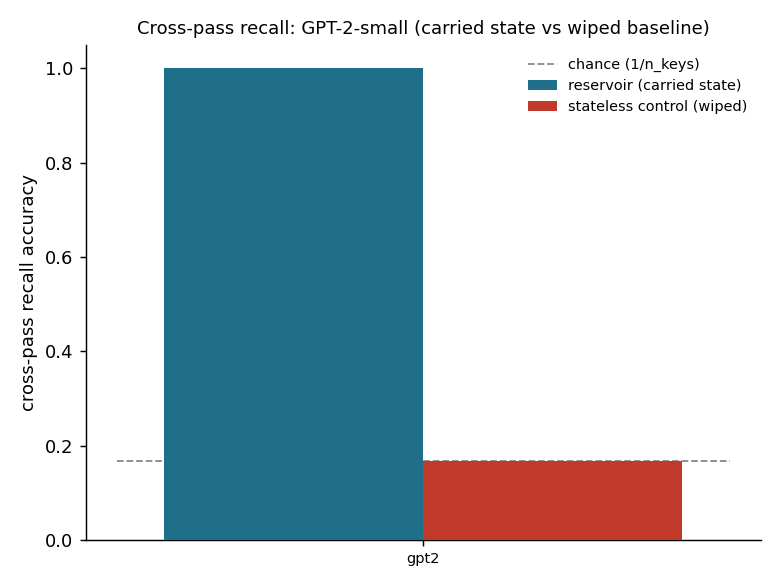}
\caption{Cross-pass recall on GPT-2-small. With the carried reservoir state the model recalls a secret word after the context is wiped (stateful), while a state-reset baseline stays at chance, the injection-design result.}
\end{figure}

\begin{figure}
\centering
\includegraphics{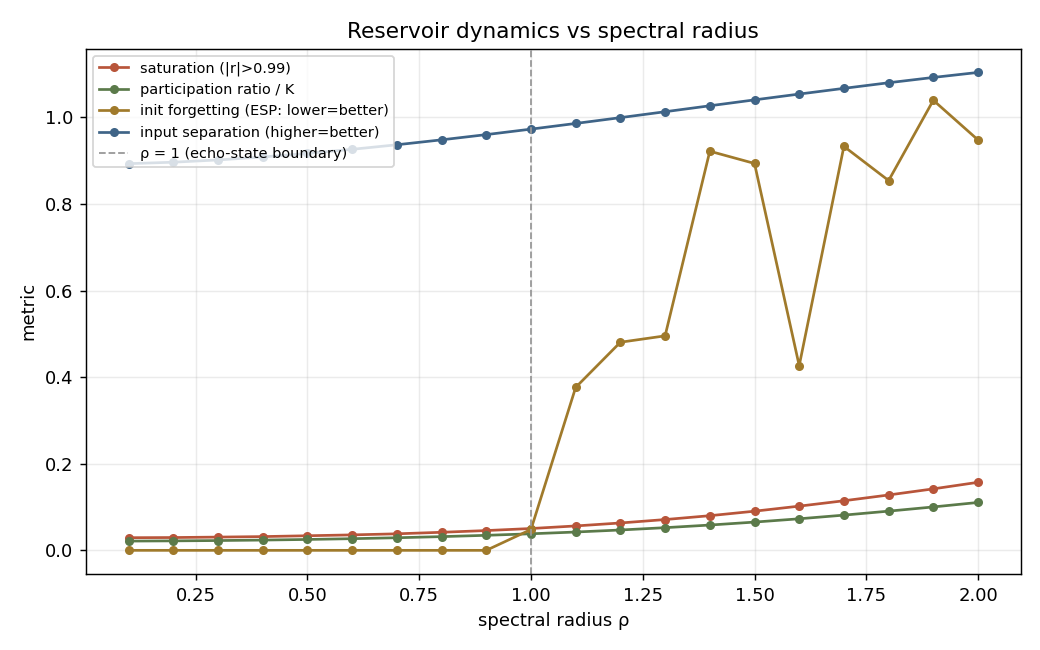}
\caption{Reservoir dynamics versus spectral radius on synthetic input. The echo state property (init-forgetting) holds cleanly below ρ ≈ 1 and breaks above it.}
\end{figure}

\begin{figure}
\centering
\includegraphics{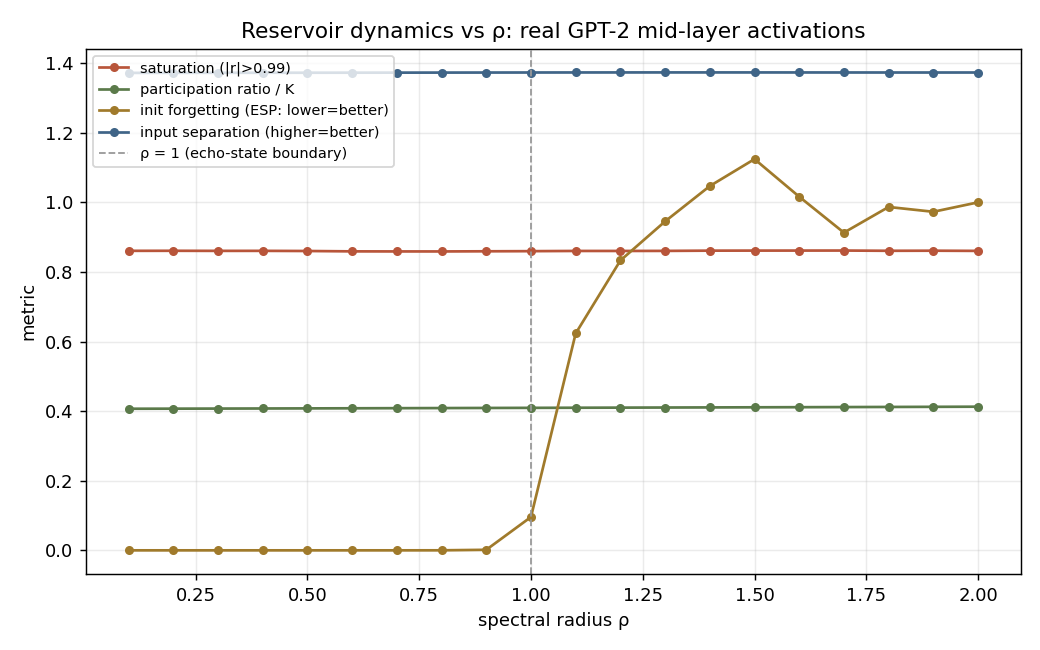}
\caption{Reservoir dynamics under real transformer activations. Saturation and effective dimensionality versus spectral radius: the regime that the unit-scaled reservoir over-drives.}
\end{figure}

\begin{figure}
\centering
\includegraphics{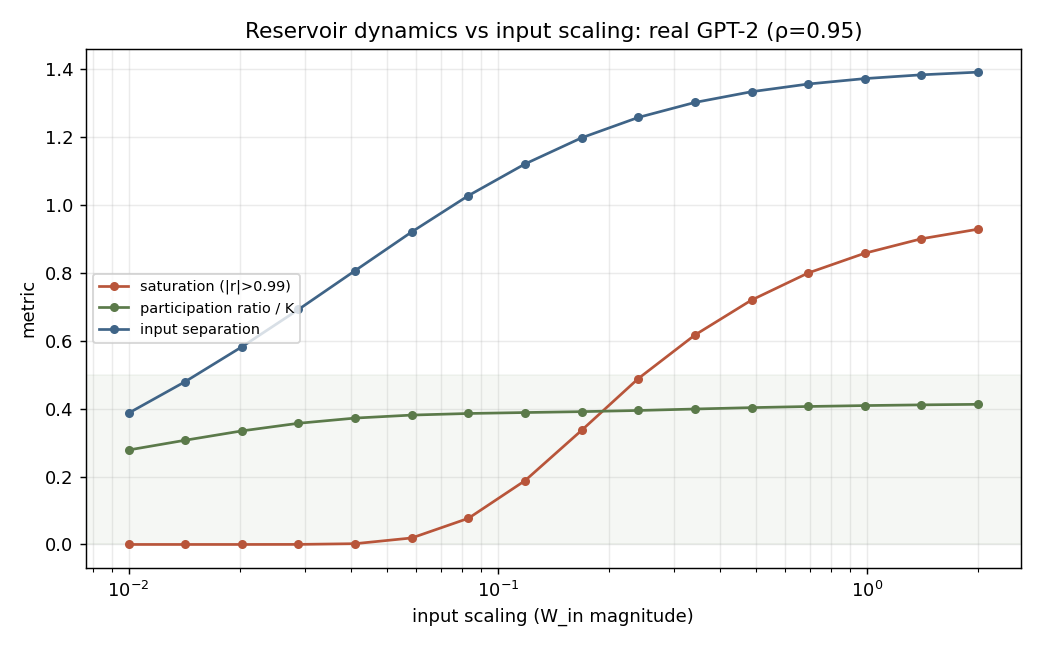}
\caption{Input-scaling sweep on real activations. The healthy band (low saturation, high usable dimensionality) sits at input scaling ≈ 0.08--0.24.}
\end{figure}

\begin{figure}
\centering
\includegraphics{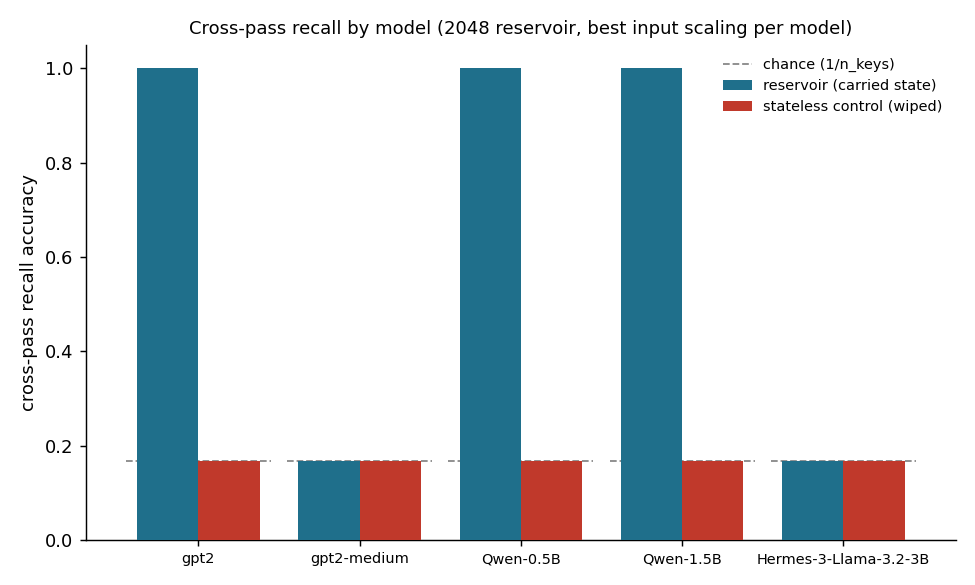}
\caption{Cross-model cross-pass recall (carried state vs wiped-reservoir control), best input scaling per model. Recall recovers at GPT-2-small and across the Qwen family (0.5B, 1.5B), but is at chance for GPT-2-medium and 4-bit Hermes-3B: not monotonic in size.}
\end{figure}

\begin{figure}
\centering
\includegraphics{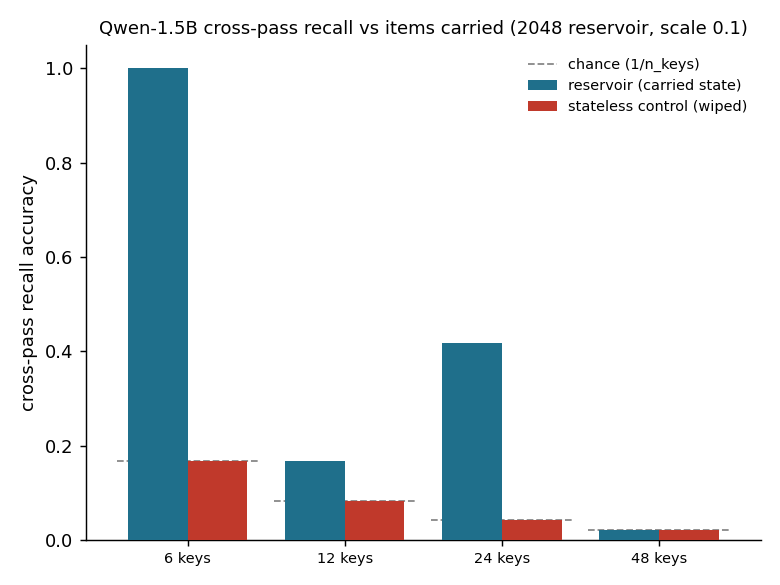}
\caption{Recall capacity at Qwen-1.5B versus the number of items carried. Perfect at six keys, strong (\textasciitilde0.42, ≈10× chance) at 24, and at chance by 48: graceful degradation into the tens of items.}
\end{figure}

\begin{figure}
\centering
\includegraphics{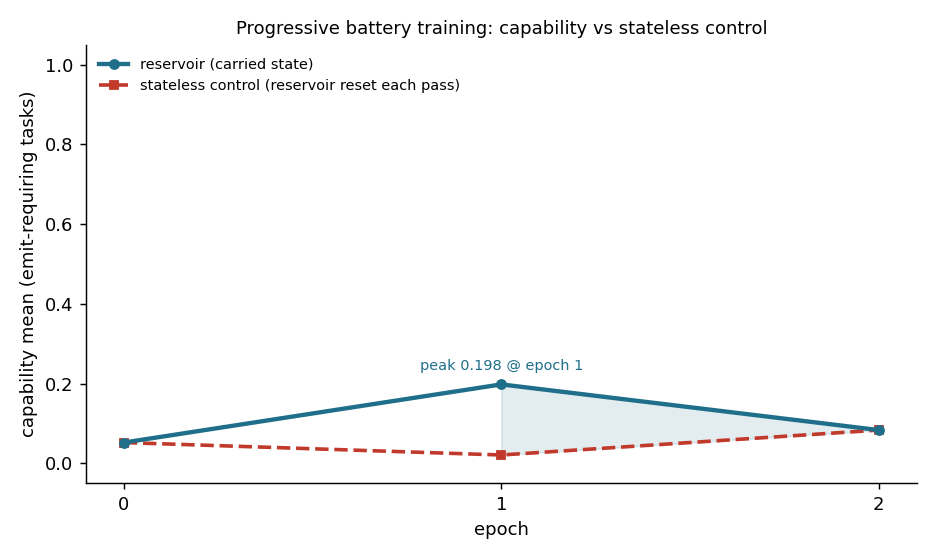}
\caption{Battery content lift versus epoch (Qwen-1.5B, content-only, eval\_n=48, chosen over eval\_n=16 so the per-epoch lift resolves above the quantization floor). The carried-state recall lift appears at epoch 1 then collapses as training drifts to a stateless solution, a within-run instance of "the model learns to ignore the recurrent state".}
\end{figure}

\begin{figure}
\centering
\includegraphics{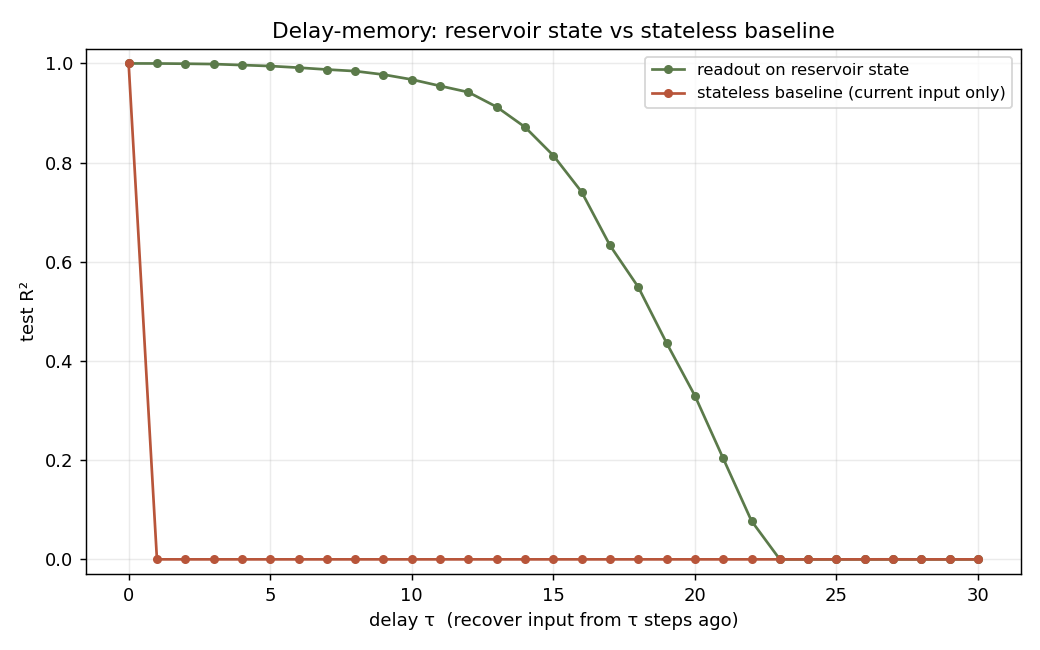}
\caption{H3 delay-memory readout. A linear readout on the reservoir state recovers delayed history that a stateless baseline provably cannot.}
\end{figure}

\begin{center}\rule{0.5\linewidth}{0.5pt}\end{center}

\emph{Reservoir Agent · research report · report site:
\url{https://reservoir.emmaleonhart.com}}

\end{document}